\documentclass{article}

\PassOptionsToPackage{numbers,compress}{natbib}

\usepackage[preprint]{neurips_2026}

\usepackage[utf8]{inputenc} 
\usepackage[T1]{fontenc}    
\usepackage{microtype}      
\usepackage{amsmath}
\usepackage{amssymb}
\usepackage{amsthm}
\usepackage{nicefrac}      
\usepackage{algorithm}
\usepackage{tabularx}
\usepackage{bbm}

\usepackage{graphicx}
\usepackage{subcaption}     
\usepackage{wrapfig}
\usepackage{booktabs}       
\usepackage{multirow}      
\usepackage{makecell}      
\usepackage[table,xcdraw]{xcolor} 
\usepackage{placeins}  
\usepackage{float}          

\usepackage{enumitem}

\usepackage{soul}

\usepackage{url}
\usepackage{hyperref}
\usepackage{longtable}

\usepackage{amsmath,amsfonts,bm}

\def\eqref#1{equation~\ref{#1}}

\def\1{\bm{1}}

\DeclareMathAlphabet{\mathsfit}{\encodingdefault}{\sfdefault}{m}{sl}
\SetMathAlphabet{\mathsfit}{bold}{\encodingdefault}{\sfdefault}{bx}{n}

\theoremstyle{plain} 
\newtheorem{theorem}{Theorem}[section]

\theoremstyle{definition}

\title{MosaicQuant: Inlier–Outlier Disaggregation for Unified 4-Bit LLM Quantization}

\author{
    Yangjia Hu\thanks{These authors contributed equally.} \\
    HKUST \\
    \texttt{huyangjia@mail.ustc.edu.cn}
    \And
    Haodong Wang\footnotemark[1] \\
    HKUST \\
    \texttt{hwanghb@connect.ust.hk}
    \And
    Zicong Hong \\
    EPFL \\
    \texttt{zicong.hong@epfl.ch}
    \AND
    Qianli Liu \\
    HKUST \\
    \texttt{qliucc@connect.ust.hk} \\
    \And
    Quanxin Shou \\
    HKUST \\
    \texttt{qshou@connect.ust.hk} \\
    \And 
    Jian Lin \\
    HKUST \\
    \texttt{jlindc@connect.ust.hk} \\
    \AND 
    Song Guo \\
    HKUST \\
    \texttt{songguo@ust.hk} \\
    \And 
    Xiaowei Shen \\
    MetaX Integrated Circuits Co., Ltd \\
    \texttt{xiaowei.shen@metax-tech.com} \\
    \And 
    Xiangjun Huang \\
    MetaX Integrated Circuits Co., Ltd \\
    \texttt{xiangjun.huang@metax-tech.com} \\
    \And 
    Dian Wang \\
    MetaX Integrated Circuits Co., Ltd \\
    \texttt{dian.wang@metax-tech.com} \\ 
    \AND 
    Jian Yang \\
    MetaX Integrated Circuits Co., Ltd \\
    \texttt{jian.yang@metax-tech.com} \\ 
  }

\begin{document}

\maketitle

\begin{abstract}
4-bit quantization significantly reduces the memory footprint and accelerates the inference of large language models (LLMs). 
However, its limited bit-width representation struggles to faithfully capture both dense common values (\emph{inliers}) and rare large-magnitude values (\emph{outliers}), causing substantial accuracy degradation. 
Existing mixed-precision methods mitigate this by retaining outliers in high precision, but at the cost of breaking the uniformity of low-bit execution, introducing precision conversion and extra data movement that undermine practical speedup.
We propose \textbf{MosaicQuant}, a unified 4-bit LLM quantization paradigm built on a novel principle of \emph{inlier--outlier disaggregation}. 
Rather than elevating outlier precision, MosaicQuant quantizes the full weight matrix into a dense 4-bit base component, where inliers are captured faithfully while outlier are inevitably quantized. 
A sparse 4-bit residual component is then introduced to compensate for these quantization errors, selectively targeting the most error-critical weight blocks where output distortion is shown to be concentrated.
However, a unified representation alone is insufficient, as naïvely executing the sparse residual as a separate kernel still breaks the unified low-bit inference pipeline. 
To bridge this gap, we introduce \textbf{ZipperEngine}, which fuses sparse block computation into the dense 4-bit GEMM kernel via an overlapped pipeline, unifying not only the representation but also the execution into a single coherent low-bit inference pipeline.
Extensive experiments on LLaMA3 and Qwen3 demonstrate that MosaicQuant preserves near-FP16 accuracy while achieving up to $1.24\times$ speedup over the W16A16 baseline.
\end{abstract}

\section{Introduction}

Large language models (LLMs) are increasingly deployed in complex applications, including code generation, autonomous agent workflows, and long-document analysis~\citep{llama3,qwen3,team2023gemini,10.1145/3586030,stcast_cvpr26,ppai_infocom26}. This remarkable progress has been driven largely by scaling model size: LLMs have grown from 7B parameters to frontier models such as DeepSeek-V3-Pro with 1.6T parameters. However, this scaling comes at a steep cost in memory footprint and computational demand, as the large-scale matrix multiplications that dominate Transformer inference become increasingly expensive, creating major bottlenecks to practical deployment.

Low-precision computation offers a practical path to alleviating these bottlenecks. Modern GPU architectures have evolved accordingly: NVIDIA's Hopper and Blackwell architectures~\citep{nvidia_blackwell} natively support low-bit integer (INT8, INT4) and floating-point (FP8, FP4) formats, enabling substantially higher matrix-multiplication throughput than traditional 16-bit execution. Quantization, which compresses LLM weights and runtime activations into these low-bit representations, has therefore become a key technique for efficient LLM inference~\citep{quantization_quarot,quantizaion_qserve,D2MoE_CPU_offload,wang2026twinquant}.

However, quantizing both weights and activations to low bit-widths often causes substantial accuracy degradation. Low-bit representations have limited dynamic range and resolution, making it difficult to faithfully capture both dense common values (\emph{inliers}) and rare large-magnitude values (\emph{outliers})~\citep{outlier}. Existing methods often address this via mixed-precision quantization~\citep{COMRT_ASPLOS,atom_quantizaiton}, preserving outlier-dominated channels or components in high precision while quantizing the remaining inlier-dominated weights to low precision. Although effective for accuracy recovery, such designs introduce high-precision fallback operations that require extra data movement and format conversion, undermining the practical hardware efficiency of low-bit inference.

\begin{figure}[t]
    \centering
    \includegraphics[width=0.97\linewidth]{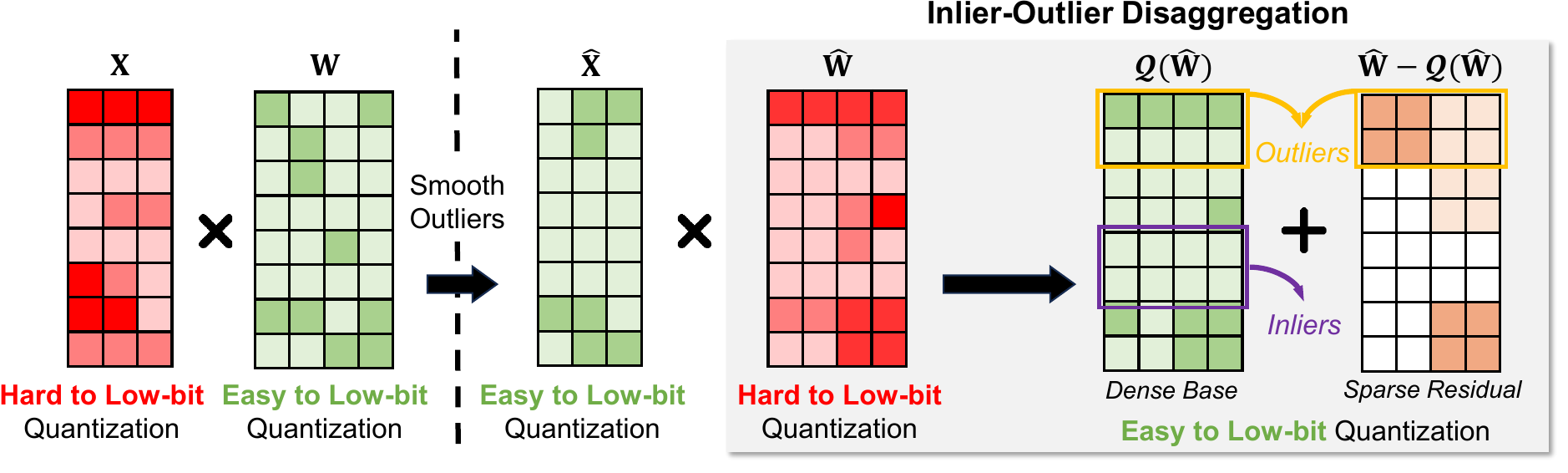}
	\caption{Overview of MosaicQuant workflow.}
    \label{overview}
\end{figure}

Therefore, we propose \textbf{MosaicQuant}, a unified 4-bit quantization paradigm that eliminates the need for high-precision fallback through a new idea of \emph{inlier--outlier disaggregation}. As shown in \autoref{overview}, rather than preserving outliers in high precision, MosaicQuant quantizes the full weight matrix into a dense 4-bit base component, where inliers are represented faithfully while outliers are inevitably quantized. A sparse 4-bit residual component is then introduced to compensate for these quantization errors. Since both components operate entirely in 4-bit, this design in principle preserves the regularity and hardware efficiency of low-bit inference, but realizing this potential requires carefully addressing two challenges in how the sparse residual is allocated and executed.

For allocation, we observe that quantization errors exhibit fine-grained block-level concentration rather than spreading uniformly across channels. Compensating entire channels would therefore include many low-impact values, whereas targeting blocks allows more precise error correction with a smaller residual budget. However, naively selecting blocks by largest absolute weight values is ineffective, as the output impact of a block also depends on the input activations. We therefore derive a Hessian-based $\Delta$-score that weights each block's residual error by activation sensitivity, enabling efficient top-\(K\) block selection without expensive per-block output evaluations.

For execution, naively computing the selected residual blocks through a separate sparse kernel does not directly translate into low latency: it requires reloading activations and merging partial outputs, adding over 30\% latency and offsetting the gains of 4-bit inference. We observe, however, that residual blocks naturally align with the tiled structure of low-bit GEMM. This allows our quantized inference engine \textbf{ZipperEngine} to fuse sparse residual computation directly into the dense kernel via an overlapped pipeline, eliminating the overhead of separate kernel dispatch and realizing the hardware efficiency that unified 4-bit representation promises. 

We summarize our contributions as follows:

\begin{itemize}[leftmargin=1.2em]
\item We propose \textbf{MosaicQuant}, a 4-bit quantization paradigm for LLMs that performs block-wise disaggregation of inlier and outlier computation, mitigating the accuracy loss of prior quantization approaches while preserving the efficiency of low-bit inference.
\item We introduce \textbf{ZipperEngine} which fuses sparse block computation into the 4-bit dense base quantization and computation kernels. This avoids the overhead of computing sparse residual blocks separately and enables practical speedup even with additional residual blocks.
\item We conduct extensive experiments on mainstream LLMs and system-level evaluations on the RTX 4090, demonstrating that MosaicQuant preserves near-FP16 accuracy while achieving up to \textbf{$1.24\times$} speedup over the W16A16 baseline.
\end{itemize}

\section{Preliminary \& Related Work}\label{relate_work}

Quantization is a widely used technique to accelerate the linear operators in Transformer layers by reducing their arithmetic precision.  
For a tensor $\mathbf{X}$, uniform symmetric quantization can be written as
\begin{equation}\label{equation 1}
  \mathbf{Q}_{\mathbf{X}} = \mathrm{round}\!\left(\frac{\mathbf{X}}{s_{\mathbf{X}}}\right),
  \quad
  s_{\mathbf{X}} = \frac{\max(|\mathbf{X}|)}{q_{\max}}, 
\end{equation}
where $\mathbf{Q}_{\mathbf{X}}$ is the low-bit representation of $\mathbf{X}$, $s_{\mathbf{X}}$ is the scaling factor, and $q_{\max}$ is the maximum representable integer. 
The dequantized tensor is then given by $\mathcal{Q}(\mathbf{X}) = s_{\mathbf{X}}\,\mathbf{Q}_{\mathbf{X}}$. For a linear layer with activation $\mathbf{X}$ and weight $\mathbf{W}$, the quantized computation can be expressed as
\begin{equation}
  \mathbf{X}\mathbf{W} \approx \mathcal{Q}(\mathbf{X})\,\mathcal{Q}(\mathbf{W})
  = s_\mathbf{X} s_\mathbf{W} \cdot \mathbf{Q}_\mathbf{X} \mathbf{Q}_\mathbf{W}.
\end{equation}
To exploit low-precision compute units on modern GPUs, $\mathbf{Q}_{\mathbf{X}}$ and $\mathbf{Q}_{\mathbf{W}}$ typically adopt the same bit-width. Otherwise, one operand must be dequantized at runtime to match the other, which undermines the performance advantage of quantized execution.

\noindent\textbf{Outlier Mitigation via Equivalent Transformations.}\quad
The input activations are often rich in outliers.
To mitigate their impact on quantization, a popular way is to apply equivalent transformations that reshape the activation and weight distributions which can be generally written as
\begin{equation}\label{equation3}
    \mathbf{\hat{X}}\mathbf{\hat{W}}
    =
    \left(\mathbf{X}\mathcal{T}\right)
    \left(\mathcal{T}^{-1}\mathbf{W}\right),
\end{equation}

where $\mathcal{T}$ denotes an invertible matrix.  
Existing methods instantiate $\mathcal{T}$ in different forms: SmoothQuant~\citep{smoothquant} uses a diagonal channel-wise scaling to migrate activation outliers into weights, while QuaRot~\citep{quantization_quarot} applies a fixed orthogonal rotation to redistribute outlier across channels. Furthermore, SpinQuant~\citep{spinquant_quant} and FlatQuant~\citep{flatquant} learn rotation matrix and affine transformation to flatten weight and activation distributions, respectively.  
These methods improve quantization by reshaping tensor distributions, but under aggressive 4-bit quantization, the remaining or migrated outliers can still induce large quantization errors in a small fraction of weight and activation blocks.

\noindent\textbf{Mixed-Precision Quantization.}\quad 
Recent studies observe that outliers in LLM weights are often concentrated in a small subset of channels or structured components~\citep{atom_quantizaiton,awq_2024}. A natural strategy is therefore to preserve these outlier-dominated parts in high-precision, while quantizing the remaining weights to low-precision~\citep{li2024svdquant,resq_ICML25,huang2025slimllm}.
For example, Atom~\citep{atom_quantizaiton} identifies outlier activation channels and keeps them in 8-bit, while applying 4-bit quantization to the normal channels. 
Moreover, SVDQuant~\citep{li2024svdquant} introduces a 16-bit low-rank branch to absorb outlier components, and quantizes the remaining residual branch to 4-bit.
Although these methods improve accuracy, they introduce additional high-precision fallback operations which require extra data movement and format conversion, limiting the practical hardware efficiency of low-bit inference. 
In contrast, MosaicQuant uses a unified sparse low-precision component to compensate outlier-induced quantization errors at the block level, preserving accuracy while maintaining a fully low-bit execution pattern.

\section{Motivation}\label{motivation}
In this paper, we first fully migrate the outliers from the activation $\hat{\mathbf{X}}$ to the weight $\hat{\mathbf{W}}$ using \autoref{equation3}. Our key insight is to represent the migrated weight with a dense 4-bit base component, while introducing an additional sparse 4-bit residual component to compensate for the quantization errors induced by outliers. Formally, we write:
\begin{equation}
\hat{\mathbf{X}}\hat{\mathbf{W}}
= \hat{\mathbf{X}}(\mathcal{Q}(\mathbf{\hat{W}})+ \mathbf{\hat{W}}-\mathcal{Q}(\mathbf{\hat{W}}))
= \hat{\mathbf{X}}\mathcal{Q}(\mathbf{\hat{W}}) + \hat{\mathbf{X}}\mathbf{R}
\approx
\underbrace{\mathcal{Q}(\mathbf{\hat{X}})\mathcal{Q}(\mathbf{\hat{W}})}_{\text{dense 4-bit component}}
+
\underbrace{\mathcal{Q}(\hat{\mathbf{X}})\mathcal{Q}(\mathbf{R})}_{\text{sparse 4-bit component}},
\end{equation}
where $\mathbf{R} = \mathbf{\hat{W}}-\mathcal{Q}(\mathbf{\hat{W}})$.
This decomposition suggests a residual-compensation path for reducing the error of the dense 4-bit base component. However, applying the residual component densely to all weights would introduce substantial extra computation and storage, weakening the efficiency advantage of low-bit inference. This motivates a sparse residual-compensation strategy, where only a small fraction of residual blocks are selected to recover most of the error-reduction effect.

\begin{figure}[t]
    \centering

    \begin{subfigure}[t]{0.49\linewidth}
        \centering
         \includegraphics[width=\linewidth]{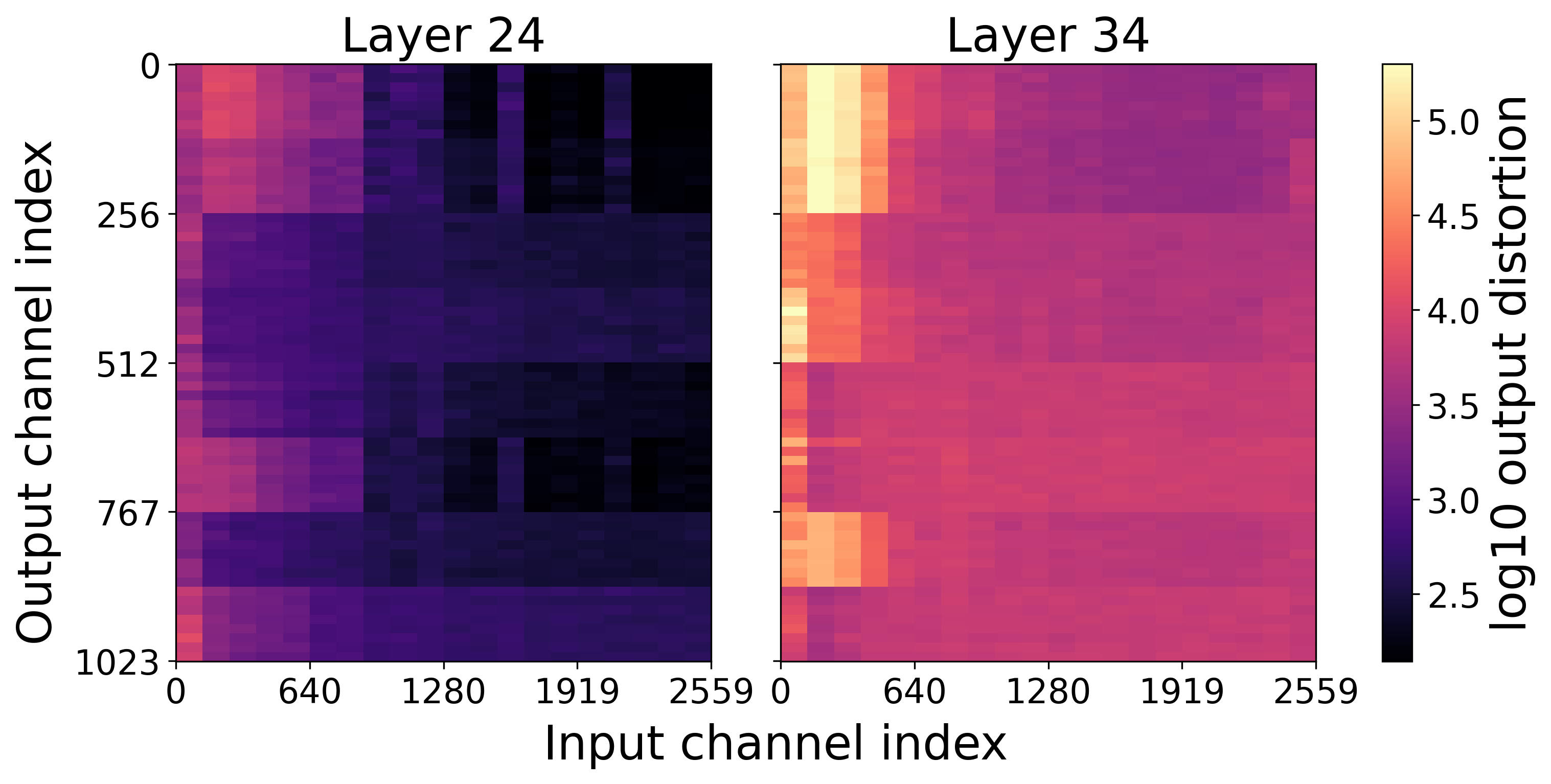}
        \caption{Block-wise concentration of residual-induced output distortion.}
        \label{fig:subfig-a}
    \end{subfigure}
    \hfill
    \begin{subfigure}[t]{0.49\linewidth}
        \centering
        \includegraphics[width=\linewidth]{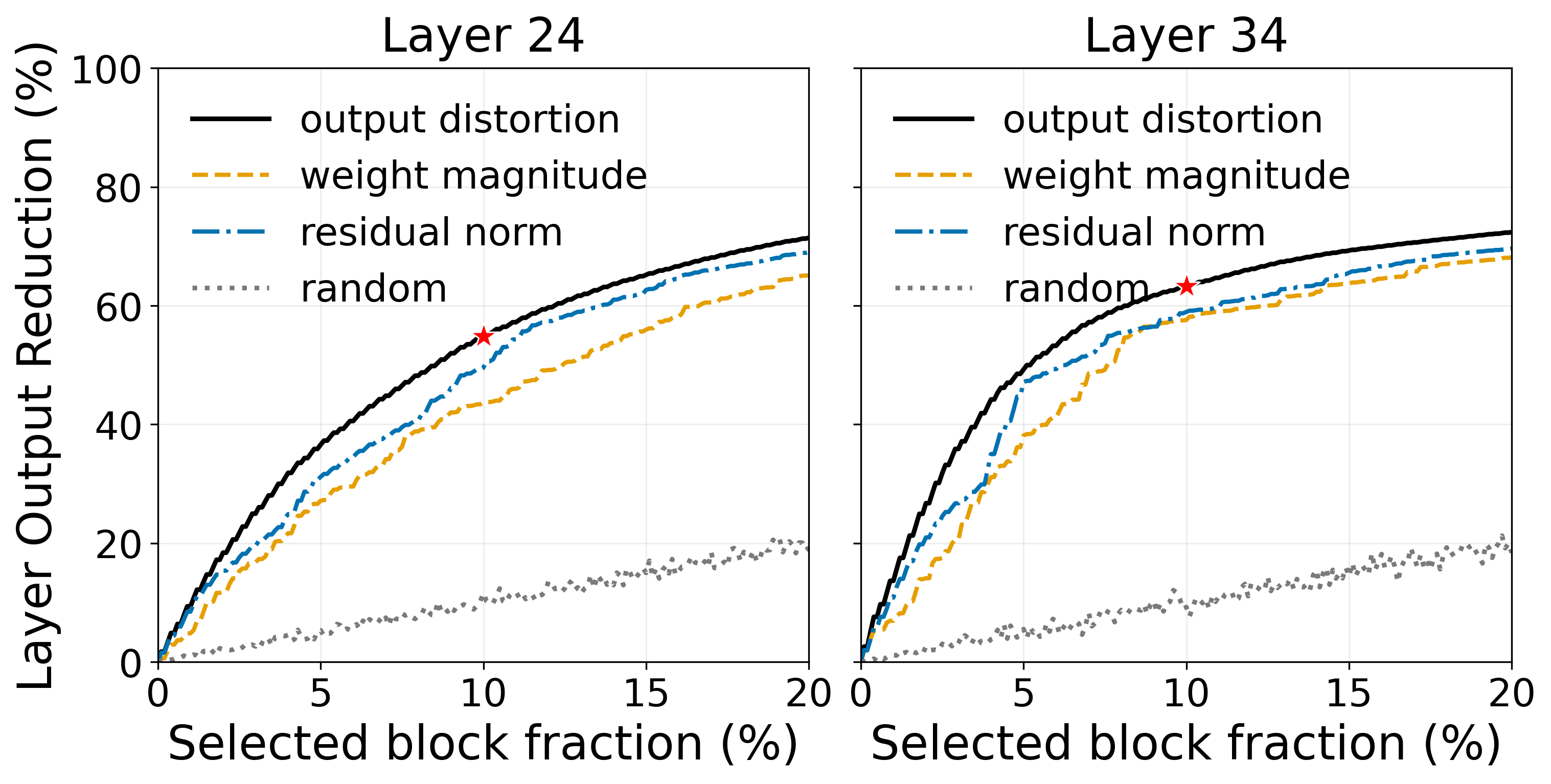}
        \caption{Comparison of block selection criteria.}
        \label{fig:subfig-b}
    \end{subfigure}

    \caption{Block-wise residual compensation analysis on the \(v_{\mathrm{proj}}\) linear layers of Qwen3-4B, including Layer 24 and Layer 34.}
    \label{fig:two-subfigures}
    \vspace{-10pt}
\end{figure}

\paragraph{Observation 1: Block-wise residual compensation is more effective than channel-wise.}
Prior studies have shown that outliers are often concentrated in a small number of channels~\citep{atom_quantizaiton,awq_2024}; thus, a naive strategy is to apply residual compensation to the corresponding weight channels.
However, we observe that the output distortion caused by residual quantization is not channel-uniform, but instead exhibits fine-grained block-level concentration.
\autoref{fig:subfig-a} visualizes the output distortion, measured by
\(\|\hat{\mathbf{X}}(\mathbf{R}-\mathcal{Q}(\mathbf{R}))\|_F^2\).
If the distortion were mainly channel-wise, high-error regions would appear as full vertical stripes in the heatmap; instead, we observe localized rectangular hotspots.
This indicates that compensating full channels would include many low-impact values, whereas block-wise compensation can target high-distortion regions more precisely.
Therefore, residual compensation should be performed at the block-level rather than the channel-level.

\paragraph{Observation 2: Output-aware selection is crucial for residual compensation.}
After establishing block-wise compensation as the basic unit, we next ask how to select the blocks to compensate.
Since residual compensation aims to preserve the layer output, the most direct criterion is the reduction in full output distortion after compensating a selected block set.
For a selected block set \(\mathcal{S}\), we measure this reduction as
\begin{equation}\label{equation5}
    C(\mathcal{S})
=
\frac{
\|\hat{\mathbf{X}}\mathbf{R}\|_F^2-\left\|
\hat{\mathbf{X}}
\left(
\mathbf{R}
-
\mathbf{M}_{\mathcal{S}} \odot \mathcal{Q}(\mathbf{R})
\right)
\right\|_F^2 
}{
\|\hat{\mathbf{X}}\mathbf{R}\|_F^2
},
\end{equation}
where \(\mathbf{M}_{\mathcal{S}}\) is a block-level binary mask that keeps \(\mathcal{Q}(\mathbf{R})\) only on the selected blocks.
\autoref{fig:subfig-b}  compares different block selection criteria under the same block budget, including output distortion selection, largest weight magnitude \(\max_{(i,j)\in B}|\hat{W}_{ij}|\), residual norm \(\|\mathbf{R}_B\|_F^2\), and random selection.
The output-distortion criterion is markedly more effective than the other selection criteria.
This advantage comes from its activation-aware nature: magnitude- and norm-based selection only measure the size of weights or residuals, ignoring their actual contribution to the layer output. 
However, directly using output distortion for block selection is expensive, because it greedily chooses the block with the largest marginal increase in \(C(\mathcal{S})\) at each step and therefore requires repeated activation-dependent full-output-distortion evaluations over all candidate blocks.
This motivates an efficient output-aware approximation for residual block selection.

\paragraph{Observation 3: A 10\% sparse residual budget recovers most quantization errors.}
The curves in \autoref{fig:subfig-b} also show that the benefit of residual compensation is highly concentrated in a small fraction of blocks.
At only a 10\% block budget, output-aware selection already attains 55\% layer output reduction on Layer 24 and 63\% on Layer 34.
In other words, compensating just 10\% blocks is sufficient to recover more than half of the output distortion.
This reveals a clear diminishing-return pattern: the most important error reduction comes from a small subset of highly critical blocks, while compensating the remaining blocks provides much smaller gains.
Therefore, only a small sparse residual budget is needed to preserve most of the accuracy benefits of residual compensation, substantially reducing the additional memory and computation overhead.

Thus, our objective is to identify a sparse set of residual blocks whose 4-bit quantized residuals can most effectively compensate for outlier-induced quantization error. Formally, we aim to find a block-wise sparse mask $\mathbf{M}$ that minimizes:
\begin{equation}\label{error_equation}
E(\hat{\mathbf{X}}, \hat{\mathbf{W}}, \mathbf{M}_{\mathcal{S}}) =
\left\lVert
\hat{\mathbf{X}}\hat{\mathbf{W}}
-
\mathcal{Q}(\hat{\mathbf{X}})
\left[
\mathcal{Q}(\hat{\mathbf{W}})
+
\mathbf{M}_{\mathcal{S}} \odot \mathcal{Q}(\mathbf{R})
\right]
\right\rVert_F^2 ,
\end{equation}
where $\mathbf{R}=\hat{\mathbf{W}}-\mathcal{Q}(\hat{\mathbf{W}})$, and $\mathbf{M}$ is a block-wise binary mask indicating which residual blocks are selected for sparse 4-bit compensation.

\section{Method}\label{method}

In this section, we present \textbf{MosaicQuant}, which preserves a unified 4-bit inference pipeline by decomposing the migrated weight into a dense 4-bit base and a sparse 4-bit residual, with output-critical residual blocks selected by an activation-aware \(\Delta\)-Hessian score. We further introduce \textbf{ZipperEngine}, a dense--sparse 4-bit execution engine that integrates these stages into an overlapped pipeline, reducing redundant memory traffic and kernel-launch overhead while improving the accuracy--efficiency trade-off.

\begin{algorithm}[t]
\caption{Hessian-Guided Residual Block Selection}
\label{alg:hessian_block_selection}
\small
\setlength{\tabcolsep}{3pt}
\renewcommand{\arraystretch}{1.08}
\begin{tabularx}{\linewidth}{@{}lX@{}}
\toprule
\multicolumn{2}{@{}l}{\textbf{Input:} migrated weight \(\hat{\mathbf{W}}\), calibration activation \(\hat{\mathbf{X}}\), block partition \(\mathcal{B}\), block budget \(K\)} \\
\multicolumn{2}{@{}l}{\textbf{Output:} selected block set \(\mathcal{S}\), block mask \(\mathbf{M}_{\mathcal{S}}\), quantized weight \(\mathbf{W}_{\mathrm{MQ}}\)} \\
\midrule
\(\mathbf{R} \leftarrow \hat{\mathbf{W}}-\mathcal{Q}(\hat{\mathbf{W}})\)
& // residual after dense 4-bit quantization \\
\(h_j \leftarrow \frac{2}{N}\sum_{n=1}^{N}\hat{\mathbf{X}}_{n,j}^{2}\)
& // activation sensitivity of input channel \(j\) \\
\textbf{for} each block \(B\in\mathcal{B}\) \textbf{do} & \\
\quad \(S_B \leftarrow
\left[
\sum_{(i,j)\in B} h_j \mathbf{R}_{ij}^{2}
-
\sum_{(i,j)\in B} h_j
\left(\mathbf{R}_{ij}-\mathcal{Q}(\mathbf{R})_{ij}\right)^{2}
\right]_+\)
& // \(\Delta\)-Hessian score \\
\textbf{end for} & \\
\(\mathcal{S} \leftarrow \operatorname{TopK}(\{S_B\}_{B\in\mathcal{B}}, K)\)
& // select residual blocks \\
\((\mathbf{M}_{\mathcal{S}})_B \leftarrow 1\) if \(B\in\mathcal{S}\), otherwise \(0\)
& // build block mask \\
\(\mathbf{W}_{\mathrm{MQ}} \leftarrow
\mathcal{Q}(\hat{\mathbf{W}})
+
\mathbf{M}_{\mathcal{S}}\odot\mathcal{Q}(\mathbf{R})\)
& // dense 4-bit base + sparse 4-bit residual \\
\bottomrule
\vspace{-10pt}
\end{tabularx}
\end{algorithm}

\subsection{MosaicQuant: Block-Wise Inlier--Outlier Disaggregation}

Building on the observations in \autoref{motivation}, MosaicQuant realizes inlier--outlier disaggregation at the block-level. 
It represents the migrated weight with a dense 4-bit base for common inlier and a sparse 4-bit residual component for compensating outlier. 
The key problem is to identify the few residual blocks that most affect the layer output, for which we derive an activation-aware \(\Delta\)-Hessian score that reduces block allocation to top-\(K\) ranking and is optimal under a diagonal-Hessian objective.

\paragraph{Hessian-Guided Block Selection.}
As discussed in \autoref{motivation}, exact output-distortion selection provides a direct objective for residual block allocation, but it is too expensive to use as a practical algorithm. 
It requires recomputing the full output distortion for each candidate block, while the effects of different blocks interact in the layer error and cannot be cheaply ranked independently.  
To retain the output-aware principle of exact selection without its computational cost, we seek a proxy metric that approximates the marginal benefit of residual compensation. 
Such a proxy should be both output-aware and block-decomposable: it should capture activation-amplified residual errors and assign each block an independent score for direct top-$K$ selection.

Inspired by GPTQ~\citep{gptq}, which uses the activation Hessian to estimate the output impact of weight quantization errors, we introduce a \(\Delta\)-Hessian score for residual block selection. 
While GPTQ uses Hessian information to reduce the error of quantizing all weights, we use it to rank which blocks deserve the limited residual 4-bit compensation. 
Specifically, for each input channel \(j\), we use the diagonal Hessian estimate \(h_j=\frac{2}{N}\sum_{n=1}^{N}\hat{X}_{n,j}^2\) to measure how strongly errors on this channel affect the layer output. 
For a block \(B\), the \(\Delta\)-Hessian score is defined as
\[
S_B
=
\left[
\sum_{(i,j)\in B} h_j \mathbf{R}_{ij}^2
-
\sum_{(i,j)\in B} h_j (\mathbf{R}_{ij}-\mathcal{Q}(\mathbf{R})_{ij})^2
\right]_+ .
\]
The score estimates the output error reduction obtained by applying residual 4-bit compensation to block \(B\), using \(h_j\) to weight how sensitive the layer output is to errors on each input channel.  
A block receives a high score only when its residual error is large and occurs on activation-sensitive channels. 
Thus, \(S_B\) favors blocks whose compensation is expected to produce the largest reduction in layer-output distortion, rather than blocks with merely large residual magnitude.

\paragraph{MosaicQuant Block Allocation.}
With the \(\Delta\)-Hessian score, residual block allocation becomes a simple score-and-sort procedure. As shown in Algorithm \autoref{alg:hessian_block_selection}, MosaicQuant computes \(S_B\) for every block, selects the top-\(K\) blocks under the sparse residual budget.
Thus, only the selected blocks receive residual 4-bit compensation, while all other blocks remain in dense 4-bit. 
This replaces expensive exact output distortion selection with one calibration-based scoring and sorting step.

\begin{figure}[t]
  \centering
  \includegraphics[width=0.9\linewidth]{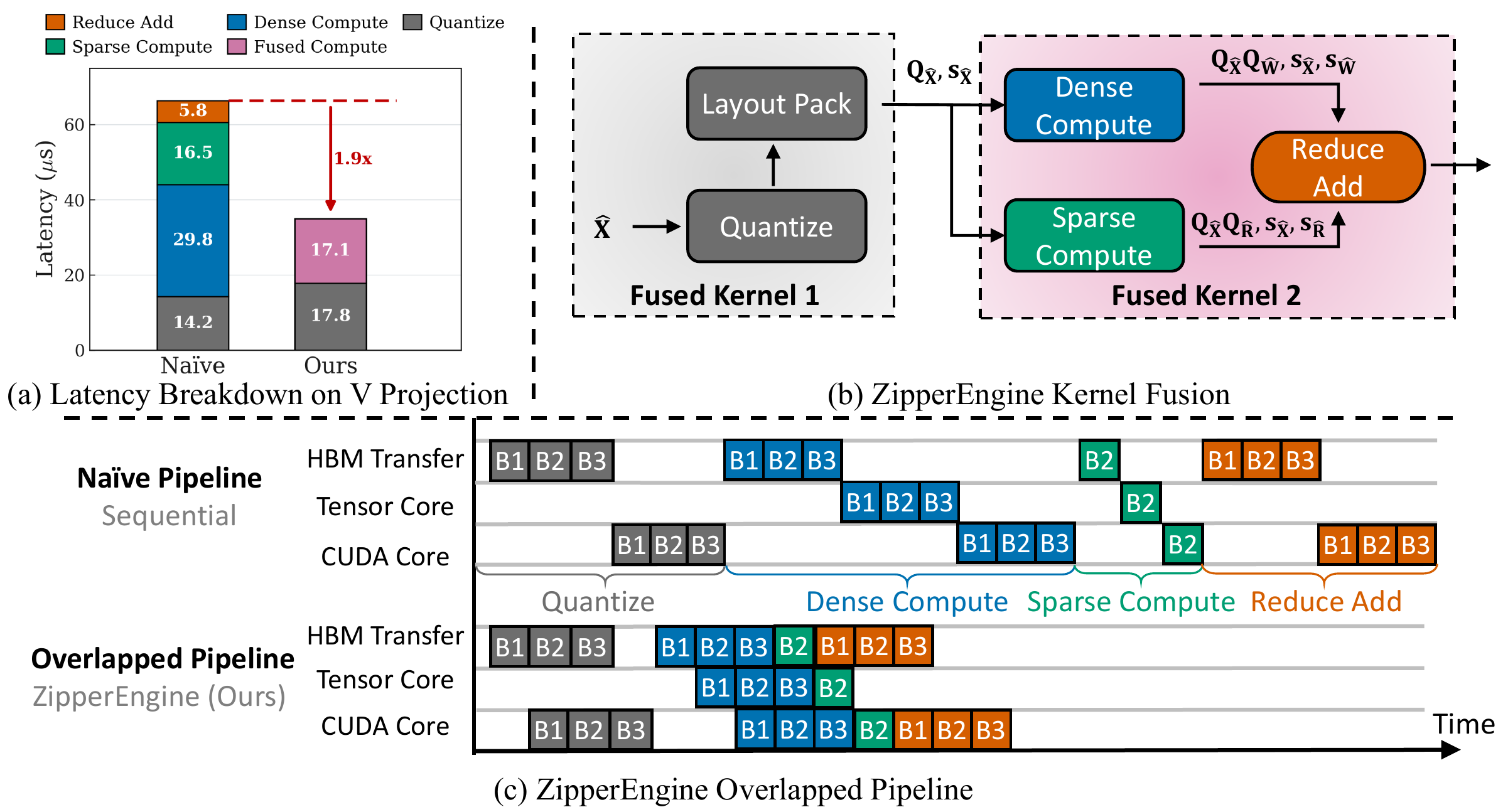}
  \caption{ZipperEngine kernel fusion and overlapped execution on the V projection layer of Qwen3-8B. By fusing dense/sparse 4-bit computation, and reduce add into an overlapped pipeline, ZipperEngine hide memory and auxiliary overheads and reduce latency by 1.9$\times$ over the naive pipeline.}
  \label{fig:zipperengine}
  \vspace{-15pt}
\end{figure}

\paragraph{Optimality under the Diagonal Hessian.}
Exact output-distortion selection is difficult because the full activation Hessian contains off-diagonal terms. 
These terms make the residual errors of different blocks interact with each other, so the benefit of selecting one block depends on which blocks have already been selected. 
Therefore, exact selection cannot be solved by ranking blocks independently.

\(\Delta\)-Hessian simplifies this problem by using only the diagonal Hessian. 
This removes the interaction terms between different blocks, while still keeping the activation sensitivity \(h_j\). 
As a result, the output-distortion objective becomes a sum of independent block-wise errors. 
Under this objective, each block has its own error-reduction gain, which is exactly \(S_B\).

\begin{theorem}[Optimality of \(\Delta\)-Hessian ranking]
Under the diagonal-Hessian objective, the optimal solution to
\[
\min_{\mathcal{S}:|\mathcal{S}|\le K}
\sum_{i,j}
h_j
\left(
\mathbf{R}_{ij}
-
(\mathbf{M}_{\mathcal{S}}\odot\mathcal{Q}(\mathbf{R}))_{ij}
\right)^2
\]
is obtained by selecting the top-\(K\) blocks with the largest \(\Delta\)-Hessian scores \(S_B\).
\end{theorem}

The proof is provided in \autoref{appendix:delta_hessian_proof}. 
This theorem shows that \(\Delta\)-Hessian is an efficient approximation to exact output-distortion selection: after removing cross-block interaction terms, it exactly optimizes the diagonal-Hessian objective. Thus, MosaicQuant preserves the output-aware principle of exact selection while reducing residual block allocation to one sorting step.

\subsection{ZipperEngine: Fusing Dense and Sparse 4-bit Kernels}\label{kernel}
ZipperEngine is designed to make MosaicQuant's sparse 4-bit residual compensation practical for low-latency inference. Its key goal is to remove the extra memory traffic and kernel-launch overhead introduced by the sparse kernels, and further overlap quantization, dense computation, sparse compensation, and reduction within a unified execution pipeline.

\paragraph{Kernel Fusion.}
We first observe that executing the sparse residual branch as a separate kernel introduces noticeable overhead. As shown in \autoref{fig:zipperengine}(a), on the V projection layer of Qwen3 8B, sparse compute takes 16.5~$\mu$s, about 55\% of the dense 4-bit compute latency. Together with reduce-add, the sparse residual path contributes 22.3~$\mu$s, nearly one-third of the naive pipeline latency. This overhead mainly comes from extra memory traffic and kernel launches: the sparse kernel reloads the quantized activation from HBM, writes a partial output, and then relies on another reduce-add kernel to merge it with the dense result.
To remove this overhead, ZipperEngine fuses the dense and sparse kernels by exploiting their shared dataflow. The sparse branch reuses the same quantized activation tiles and accumulates into the same output tiles as the dense GEMM, as shown in \autoref{fig:zipperengine}(b). ZipperEngine therefore fuses layout packing with quantization, and further integrates sparse residual computation and reduce-add into the dense GEMM kernel. This removes redundant HBM accesses and kernel launches, making sparse residual compensation nearly cost-free.

\begin{table*}[t]
\caption{Comparison of the perplexity score on WikiText2 and averaged accuracy on six zero-shot commonsense reasoning tasks.}
\label{precision}
\centering
\scriptsize
\setlength{\tabcolsep}{3.2pt}
\renewcommand{\arraystretch}{1.08}
\begin{tabular}{@{}ll
cc@{\hspace{6pt}}
cc@{\hspace{6pt}}
cc@{\hspace{6pt}}
cc@{\hspace{6pt}}
cc@{\hspace{6pt}}
cc@{}
}
\toprule
\multirow{3}{*}{\textbf{Precision}} &
\multirow{3}{*}{\textbf{Method}} &
\multicolumn{2}{c}{\textbf{LLaMA3.2 3B}} &
\multicolumn{2}{c}{\textbf{LLaMA3 8B}} &
\multicolumn{2}{c}{\textbf{Qwen3 4B}} &
\multicolumn{2}{c}{\textbf{Qwen3 8B}} &
\multicolumn{2}{c}{\textbf{Qwen3 14B}} &
\multicolumn{2}{c}{\textbf{Qwen3 32B}} \\
\cmidrule(lr){3-4}\cmidrule(lr){5-6}\cmidrule(lr){7-8}\cmidrule(lr){9-10}\cmidrule(lr){11-12}\cmidrule(lr){13-14}
& &
\makecell{0-shot\\Avg.} & \makecell{Wiki\\($\downarrow$)} &
\makecell{0-shot\\Avg.} & \makecell{Wiki\\($\downarrow$)} &
\makecell{0-shot\\Avg.} & \makecell{Wiki\\($\downarrow$)} &
\makecell{0-shot\\Avg.} & \makecell{Wiki\\($\downarrow$)} &
\makecell{0-shot\\Avg.} & \makecell{Wiki\\($\downarrow$)} &
\makecell{0-shot\\Avg.} & \makecell{Wiki\\($\downarrow$)} \\
\midrule
W16A16 & -- & 66.2 & 10.7 & 73.5 & 8.6 & 69.4 & 10.0 & 71.6 & 9.7 & 74.2 & 8.6 & 75.2 & 7.6\\
\midrule
\multirow{3}{*}{W4A16}
    & RTN  & 60.9 & 18.8 & 71.2 & 10.5 & 67.1 & 10.6 & 69.0 & 12.0 & 70.3 & 9.9 & 68.1 & 38.5\\
    & GPTQ & 61.7 & 15.2 & 72.4 & 9.0  & 68.6 & 10.4 & 70.4 & 10.8 & 72.7 & 9.2 & 73.5 & 8.3 \\
    & AWQ  & 63.1 & 12.7 & 72.6 & 10.3 & 68.4 & 10.4 & 70.8 & 10.2 & 69.8 & 9.6 & 73.8 & 8.2 \\
\midrule
\multirow{7}{*}{W4A8}
    & RTN          & 60.7 & 29.0  & 71.0 & 10.6 & 58.8 & 30.6 & 64.4 & 12.3 & 68.3 & 10.9 & 67.4 & 11.2\\
    & SmoothQuant  & 59.8 & 288.5 & 60.4 & 13.3 & 59.9 & 22.6 & 63.8 & 12.5 & 68.1 & 11.0 & 67.1 & 11.6\\
    & QuaRot       & 64.2 & 12.4  & 71.6 & 10.6 & 67.8 & 11.2 & 67.1 & 11.5 & 70.7 & 10.4 & 71.0 & 10.5\\
    & SpinQuant    & 65.2 & 11.5  & 72.3 & 10.3 & 68.4 & 10.8 & 70.4 & 10.8 & 72.5 & 9.8  & 73.1 & 8.9 \\
    & ResQ         & 64.8 & 11.8  & 72.2 & 10.4 & 68.4 & 10.7 & 70.3 & 10.7 & 72.4 & 12.1 & 73.0 & 9.2 \\
    & Atom         & 64.5 & 11.9  & 71.9 & 10.6 & 67.8 & 11.0 & 70.3 & 10.9 & 72.1 & 12.4 & 73.0 & 9.3 \\
    & \cellcolor{gray}MosaicQuant
                  & \cellcolor{gray}65.3 & \cellcolor{gray}11.3
                  & \cellcolor{gray}72.7 & \cellcolor{gray}10.2
                  & \cellcolor{gray}68.7 & \cellcolor{gray}10.4
                  & \cellcolor{gray}70.6 & \cellcolor{gray}10.6
                  & \cellcolor{gray}73.4 & \cellcolor{gray}11.5
                  & \cellcolor{gray}74.2 & \cellcolor{gray}8.4 \\
\midrule
\multirow{7}{*}{W4A4}
    & RTN          & 43.2 & 741.9 & 40.4 & 92.9  & 41.5 & 8791  & 40.1 & 4392   & 42.1 & 18749 & 45.4 & 1796\\
    & SmoothQuant  & 44.7 & 372.3 & 52.8 & 19.4  & 40.1 & 9910  & 40.1 & 3360.1 & 41.3 & 21675 & 44.4 & 1806\\
    & QuaRot       & 61.4 & 16.9  & 68.0 & 13.1  & 61.7 & 12.5  & 66.3 & 15.6   & 67.2 & 18.2  & 67.4 & 15.6\\
    & SpinQuant    & 64.1 & 11.9  & 70.0 & 11.4  & 65.6 & 11.5  & 68.3 & 11.4   & 71.7 & 12.0  & 72.7 & 10.3\\
    & ResQ         & 63.0 & 12.2  & 69.9 & 11.6  & 65.1 & 12.0  & 66.8 & 14.2   & 71.0 & 12.8  & 72.0 & 10.9\\
    & Atom         & 62.6 & 12.5  & 67.4 & 12.5  & 64.8 & 12.3  & 66.1 & 14.5   & 70.5 & 14.5  & 72.2 & 10.5\\
    & \cellcolor{gray}MosaicQuant
                  & \cellcolor{gray}64.9 & \cellcolor{gray}11.8
                  & \cellcolor{gray}71.2 & \cellcolor{gray}10.8
                  & \cellcolor{gray}66.3 & \cellcolor{gray}11.0
                  & \cellcolor{gray}68.6 & \cellcolor{gray}11.2
                  & \cellcolor{gray}72.9 & \cellcolor{gray}11.8
                  & \cellcolor{gray}73.3 & \cellcolor{gray}8.8 \\
\bottomrule
\vspace{-10pt}
\end{tabular}
\end{table*}

\paragraph{Overlapped pipeline.}
Kernel fusion removes redundant memory traffic, but the fused stages can still execute sequentially. As shown in \autoref{fig:zipperengine}(c), quantization, dense compute, sparse compute, and reduce-add are performed one after another, leaving HBM, Tensor Cores, and CUDA Cores idle at different times.
To improve utilization, ZipperEngine adopts a tiled overlapped pipeline. It splits each tile into independent sub-blocks and schedules different stages concurrently. For example, while Tensor Cores execute dense MMA for sub-block B1, CUDA Cores can prepare quantization and layout packing for B2, and HBM can prefetch sparse residual blocks for B3. Since each sub-block reads its own input tiles and accumulates into its own output tile, these stages do not introduce cross-tile dependencies. Reduce-add is also folded into the local fused output path instead of launched as a separate kernel. This software pipeline hides auxiliary work behind dense 4-bit GEMM, keeping hardware units busy while preserving the throughput of regular 4-bit inference.

\section{Experiments}

\subsection{Setups}
\label{section5-1}
\textbf{Models and metric.} We benchmark against two mainstream LLMs: the LLaMA3~\citep{llama3} models (3B/8B) and Qwen3~\citep{qwen3} models (4B/8B/14B/32B). Following previous works~\citep{flatquant,spinquant_quant}, we randomly sample the 128 prompts from WikiText-2 dataset~\citep{wiki_dataset} for calibration, each sentence with 2048 tokens. To evaluate the commonsense reasoning capability of our method, we use six zero-shot evaluation tasks, including ARC-Challenge, ARC-Easy~\citep{ARC-C}, HellaSwag~\citep{hellaswag}, LAMBADA~\citep{lambada}, PIQA\citep{PIQA}, and WinoGrande~\citep{WinoGrande}. Additionally, we also report the perplexity score on WikiText2~\citep{wiki_dataset} datasets. All accuracy metrics are tested on lm-eval~\citep{eval-harness}. We also evaluate the memory save and speedup.

\textbf{Baselines.}
We compare MosaicQuant against six popular PTQ methods, including the naive quantization method RTN~\citep{RTN}, the weight-only quantization GPTQ~\citep{gptq} and AWQ~\citep{awq_2024}, the mixed-precision quantizaiton ResQ~\citep{resq_ICML25} and Atom~\citep{atom_quantizaiton}, the weight-activation quantization SmoothQuant~\citep{smoothquant} and two recent state-of-the-art methods QuaRot~\citep{quantization_quarot}, SpinQuant~\citep{spinquant_quant}. See Appendix~\ref{baseline} for more details.

\textbf{Implementation details.} 
Please refer to Appendix \ref{implementation} fore more details.

\subsection{Accuracy results}
\autoref{precision} presents the comparison of WikiText2 perplexity and average accuracy on six zero-shot tasks for LLaMA3 and Qwen3 models. See more details in Appendix \ref{zero_shot_appendix}.

\textbf{Language generation tasks.}
Under W4A8, MosaicQuant achieves the lowest WikiText2 perplexity among the quantized methods, with only a small gap to W16A16. On LLaMA3.2-3B, for example, it lowers the SpinQuant perplexity from 11.5 to 11.3, while W16A16 is 10.7. It is also better than the mixed-precision baselines that keep outlier channels in 8 bits. On Qwen3-32B, ResQ and Atom give perplexities of 9.2 and 9.3, whereas MosaicQuant reaches 8.4. This suggests that spending the extra precision on selected residual blocks is more effective than keeping whole outlier channels in 8 bits.

The gap is clearer at W4A4. RTN and SmoothQuant often collapse on Qwen3 models, with perplexity reaching ~$10^3$. QuaRot and Atom are more stable, but still degrade noticeably. MosaicQuant avoids these collapses and keeps Qwen3 perplexity between 8.8 and 11.8. Against the stronger W4A4 baselines, the gains are smaller but consistent: 0.1--1.5 points over SpinQuant. The largest gap to the mixed-precision baselines appears on Qwen3-8B, where MosaicQuant reaches 11.2 perplexity, compared with 14.2 for ResQ and 14.5 for Atom.

\textbf{Zero-shot tasks.}
The same pattern appears in zero-shot accuracy, though with smaller margins. Under W4A8, MosaicQuant is close to the W4A16 baselines and exceeds them on several models. For example, it is 2.2 points above AWQ on LLaMA3.2-3B and 3.6 points above AWQ on Qwen3-14B. It also stays ahead of ResQ and Atom, with margins ranging from 0.3 to 1.3 points.

Under W4A4, MosaicQuant preserves more zero-shot accuracy than the other low-bit baselines. On LLaMA3-8B and Qwen3-14B, it is 1.3--3.8 points higher than ResQ and Atom, and across the Qwen family it is 0.3--1.2 points higher than SpinQuant. The remaining loss to W16A16 is 1.3--3.1 percentage points across the six models, smaller than for the other W4A4 baselines.

\begin{figure}[t]
    \centering
    \vspace{-15pt}
    \includegraphics[width=0.8\linewidth]{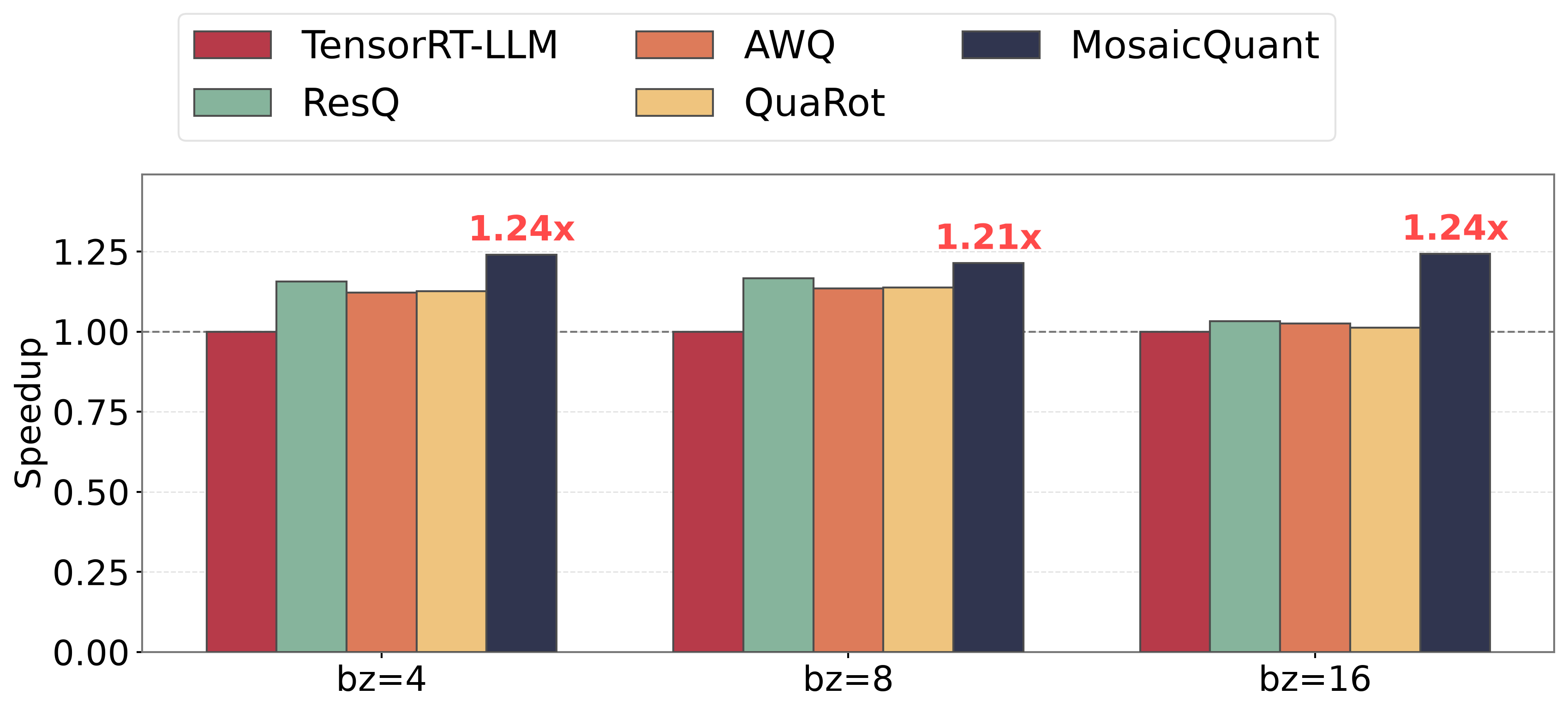}
	\caption{Throughput speedup comparison on Qwen3-8B with input length 2048 and output length 128. MosaicQuant consistently achieves the highest throughput across batch sizes 4 to 32.}
\vspace{-15pt}
\label{fig:speedup}
\end{figure}

\subsection{End-to-end Throughput}

\autoref{fig:speedup} compares MosaicQuant with TensorRT-LLM and recent quantization baselines on Qwen3-8B, using an input length of 2048 and an output length of 128. MosaicQuant consistently delivers the highest throughput across batch sizes 4--32. Compared with ResQ, the strongest competing baseline, MosaicQuant achieves 1.11--1.20$\times$ higher throughput. This gain comes from the unified dense--sparse INT4 execution: the dense 4-bit base and sparse 4-bit residual compensation are computed in a coordinated pipeline, so the compensation branch introduces limited extra memory traffic and can be overlapped with the main low-bit computation. In contrast, ResQ also aims to recover quantization errors, but its residual recovery path is less tightly integrated with the main GEMM pipeline, leading to additional scheduling and memory-access overhead.

MosaicQuant also outperforms AWQ by 1.32--1.54$\times$. Although AWQ reduces weight memory through weight-only quantization, its activations remain in high-precision, so the inference path still suffers from larger activation movement and high-bit compute utilization. Compared with QuaRot, MosaicQuant obtains 1.26--1.37$\times$ higher throughput because QuaRot relies on rotation-based outlier redistribution, which can introduce extra transformation before low-bit computation. By contrast, MosaicQuant directly compensates only the output-critical residual blocks and keeps both the dense and sparse branches in INT4, avoiding high-precision fallback while preserving a hardware-friendly low-bit pipeline.
Additional kernel profiling results are reported in Appendix~\ref{kernel_profile}.

\subsection{Ablation study}
\begin{figure*}[t]
    \centering
    \setlength{\tabcolsep}{0pt}

    \newlength{\captionblockheight}
    \setlength{\captionblockheight}{38pt}

    \begin{tabular}{@{}p{0.46\textwidth}@{\hspace{0.015\textwidth}}p{0.525\textwidth}@{}}

    \begin{minipage}[t]{\linewidth}
        \begin{minipage}[t][\captionblockheight][t]{\linewidth}
            \captionsetup{type=table,position=top,skip=0pt}
            \caption{Ablation study of MosaicQuant on Qwen3 4B and LLaMA3 8B, reporting the average accuracy on six zero-shot tasks and WikiText-2 perplexity.}
            \label{ablation_accuracy}
        \end{minipage}
        \vspace{2pt}

        \centering
        \scriptsize
        \setlength{\tabcolsep}{1.5pt}
        \renewcommand{\arraystretch}{0.95}
        \begin{tabular}{cccccc}
            \toprule
            \multirow{3}{*}{\textbf{Precision}} & \multirow{3}{*}{\textbf{Method}} & \multicolumn{2}{@{}c@{}}{\textbf{Qwen3 4B}} & \multicolumn{2}{@{}c@{}}{\textbf{LLaMA3 8B}} \\
            \cmidrule(lr){3-4}\cmidrule(lr){5-6}
            & & \makecell{0-shot\\Avg.} & \makecell{Wiki\\($\downarrow$)} & \makecell{0-shot\\Avg.} & \makecell{Wiki\\($\downarrow$)} \\
            \midrule
            W16A16 & -- & 69.4 & 10.0 & 73.5 & 8.6 \\
            \midrule
            \multirow{5}{*}{W4A4} 
                                  & w/o Sparse Branch   & 52.4 & 108.6 & 55.7 & 143.7 \\
                                  & w/ Channel Compensation     & 62.6 & 13.5 & 69.8 & 13.1 \\
                                  & MosaicQuant    & 66.3 & 11.0  & 71.2 & 10.8 \\
            \bottomrule
        \end{tabular}
    \end{minipage}
    &

    \begin{minipage}[t]{\linewidth}
        \begin{minipage}[t][\captionblockheight][t]{\linewidth}
            \captionsetup{type=figure,position=top,skip=0pt}
            \caption{Memory and latency overhead of sparse 4-bit branches across six models under 10\% block budget.}
            \label{fig:overhead}
        \end{minipage}

        \centering
        \includegraphics[width=\linewidth]{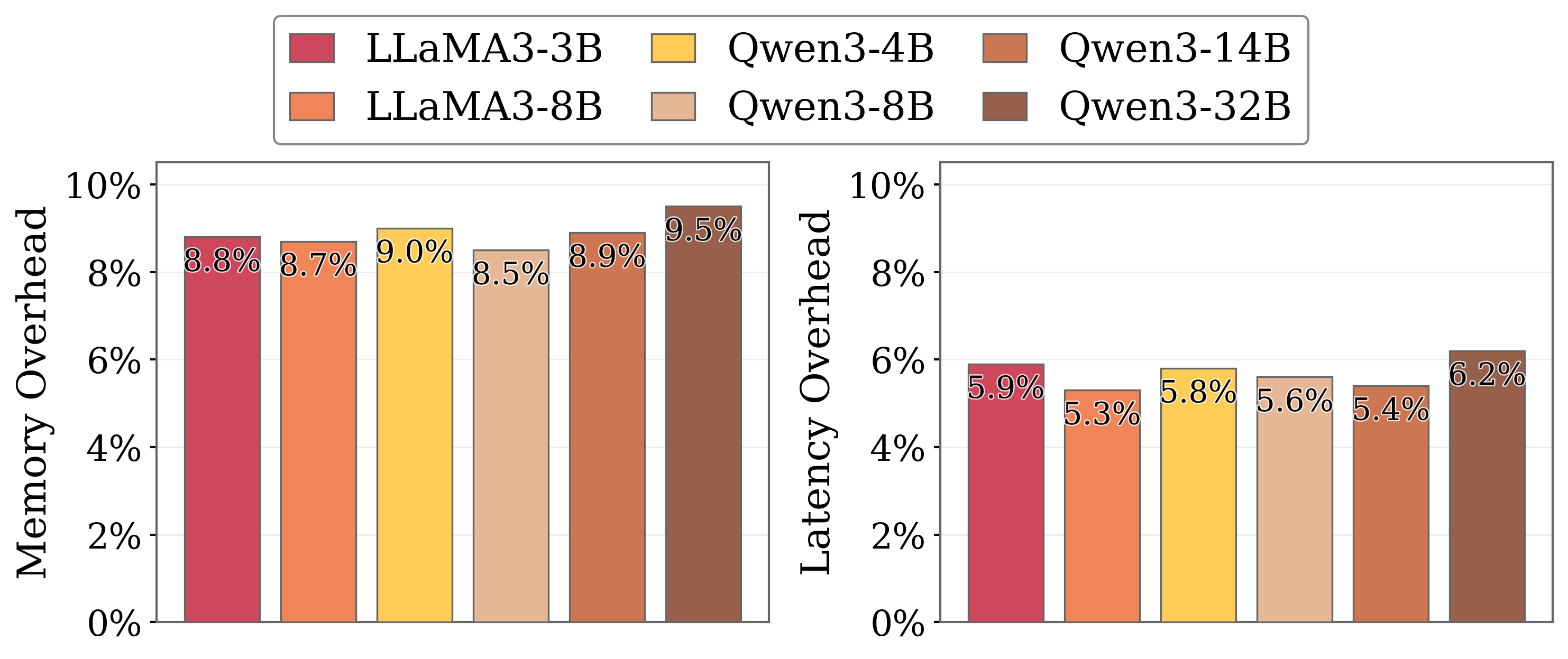}
    \end{minipage}

    \end{tabular}
    \vspace{-15pt}
\end{figure*}

As illustrated in \autoref{ablation_accuracy}, we ablate MosaicQuant on Qwen3 4B and LLaMA3 8B under W4A4 quantization. 
Removing the sparse residual branch causes severe degradation, reducing zero-shot accuracy to 52.4 on Qwen3 4B and 55.7 on LLaMA3 8B, while increasing WikiText-2 perplexity to 108.6 and 143.7, respectively. 
Adding a 10\% channel-level residual budget substantially improves accuracy to 62.6 and 69.8, and reduces perplexity to 13.5 and 13.1, confirming the importance of residual compensation. 
Replacing coarse channel-level compensation with our Hessian-guided block selection further improves accuracy to 66.3 and 71.2, and lowers perplexity to 11.0 and 10.8. 
These results demonstrate that fine-grained, output-aware block selection allocates the residual budget more effectively than channel-level compensation.

\subsection{Overhead of Sparse 4-bit Branches}
As shown in \autoref{fig:overhead}, sparse 4-bit residual branches incur only modest overhead across all six models. The memory overhead stays within 8.5--9.5\%, since only a small fraction of selected blocks receive an additional 4-bit residual branch instead of dense or high-precision compensation. Meanwhile, the latency overhead remains lower, ranging from 5.3--6.2\%, indicating that sparse compensation does not translate into proportional runtime cost. This is enabled by our fused overlapped pipeline, which co-schedules the dense 4-bit base path and sparse 4-bit residual path to reduce extra memory traffic and kernel-launch overhead.

\section{Conclusion}

In this work, we introduce \textbf{MosaicQuant}, a 4-bit post-training quantization paradigm for LLMs. It performs block-level inlier--outlier disaggregation, representing weights with a dense 4-bit base and a sparse 4-bit residual selected by a Hessian-guided output-aware criterion to compensate outlier-containing blocks that dominate layer-output error. \textbf{ZipperEngine} further fuses the sparse residual into dense 4-bit GEMM and pipelines quantization, compute, and reduce-add into one overlapped kernel, eliminating extra memory traffic from sparse compensation. Experiments show that MosaicQuant preserves near-FP16 accuracy, while ZipperEngine achieves up to $1.24\times$ speedup over the 16-bit baseline on an NVIDIA RTX~4090 GPU, enabling efficient LLM serving without sacrificing the regularity of uniform 4-bit execution.

\bibliography{reference}
\bibliographystyle{plainnat}

\appendix

\section{Proof of \(\Delta\)-Hessian Block Selection}
\label{appendix:delta_hessian_proof}

We prove that the proposed \(\Delta\)-Hessian score gives the optimal block
selection rule under a diagonal Hessian surrogate of the exact output-distortion
objective. We then show how this surrogate approximates the exact
output-distortion oracle.

Let \(\mathbf{R}\) be the residual after dense 4-bit quantization, and let
\(\mathcal{Q}(\mathbf{R})\) be the 4-bit residual compensation candidate. For a
selected block set \(\mathcal{S}\), let \(\mathbf{M}_{\mathcal{S}}\) be the
corresponding block-wise binary mask. The remaining residual error after
compensation is
\[
\mathbf{E}_{\mathcal{S}}
=
\mathbf{R}
-
\mathbf{M}_{\mathcal{S}}\odot \mathcal{Q}(\mathbf{R}).
\]
The exact layer-output distortion is
\[
L(\mathcal{S})
=
\frac{2}{N}
\left\|
\hat{\mathbf{X}}\mathbf{E}_{\mathcal{S}}^\top
\right\|_F^2,
\]
where the constant factor \(\frac{2}{N}\) does not affect block selection. Let
\[
\mathbf{H}
=
\frac{2}{N}
\hat{\mathbf{X}}^\top \hat{\mathbf{X}}
\]
be the activation Hessian. Then
\[
L(\mathcal{S})
=
\operatorname{Tr}
\left(
\mathbf{E}_{\mathcal{S}}
\mathbf{H}
\mathbf{E}_{\mathcal{S}}^\top
\right).
\]
The exact output-distortion oracle selects blocks by maximizing
\[
\Delta(\mathcal{S})
=
L(\emptyset)-L(\mathcal{S}),
\]
where \(L(\emptyset)\) is the output distortion without residual compensation.

However, the full Hessian contains cross-channel interactions. Expanding the
trace gives
\[
L(\mathcal{S})
=
\sum_i
\sum_{j,k}
H_{j,k}
(\mathbf{E}_{\mathcal{S}})_{i,j}
(\mathbf{E}_{\mathcal{S}})_{i,k}.
\]
When \(j\neq k\), the term \(H_{j,k}\) couples the errors from different input
channels. As a result, the contribution of one block depends on the errors left
in other blocks, so the exact objective cannot be decomposed into independent
block scores.

To obtain a decomposable surrogate, we approximate the Hessian by its diagonal:
\[
\mathbf{D}
=
\operatorname{diag}(h_1,\ldots,h_{d_{\mathrm{in}}}),
\qquad
h_j
=
\frac{2}{N}
\sum_{n=1}^{N}
\hat{X}_{n,j}^2.
\]
This gives the surrogate distortion
\[
\widetilde{L}(\mathcal{S})
=
\operatorname{Tr}
\left(
\mathbf{E}_{\mathcal{S}}
\mathbf{D}
\mathbf{E}_{\mathcal{S}}^\top
\right)
=
\sum_{i,j}
h_j
\left(
\mathbf{R}_{ij}
-
(\mathbf{M}_{\mathcal{S}}\odot \mathcal{Q}(\mathbf{R}))_{ij}
\right)^2.
\]

For a block \(B\), define
\[
A_B
=
\sum_{(i,j)\in B}
h_j \mathbf{R}_{ij}^2
\]
as the surrogate error contributed by block \(B\) without residual compensation,
and
\[
B_B
=
\sum_{(i,j)\in B}
h_j
\left(
\mathbf{R}_{ij}
-
\mathcal{Q}(\mathbf{R})_{ij}
\right)^2
\]
as the surrogate error contributed by block \(B\) after applying residual
4-bit compensation. The error reduction of selecting block \(B\) is
\[
\delta_B
=
A_B-B_B.
\]
This is exactly the unclipped \(\Delta\)-Hessian score:
\[
\delta_B
=
\sum_{(i,j)\in B}
h_j \mathbf{R}_{ij}^2
-
\sum_{(i,j)\in B}
h_j
\left(
\mathbf{R}_{ij}
-
\mathcal{Q}(\mathbf{R})_{ij}
\right)^2.
\]
The score used in the algorithm is
\[
S_B=[\delta_B]_+.
\]

Now we show that top-\(K\) selection is optimal for the surrogate objective.
Because the blocks form a disjoint partition of the weight matrix, the surrogate
distortion can be decomposed as
\[
\widetilde{L}(\mathcal{S})
=
\sum_{B\notin \mathcal{S}} A_B
+
\sum_{B\in \mathcal{S}} B_B.
\]
Equivalently,
\[
\widetilde{L}(\mathcal{S})
=
\sum_{B} A_B
-
\sum_{B\in \mathcal{S}}(A_B-B_B)
=
\widetilde{L}(\emptyset)
-
\sum_{B\in \mathcal{S}}\delta_B.
\]
Since \(\widetilde{L}(\emptyset)\) is constant with respect to
\(\mathcal{S}\), minimizing \(\widetilde{L}(\mathcal{S})\) under the budget
constraint \(|\mathcal{S}|\le K\) is equivalent to
\[
\max_{\mathcal{S}}
\sum_{B\in \mathcal{S}} \delta_B,
\qquad
\mathrm{s.t.}\quad
|\mathcal{S}|\le K.
\]
This is a modular maximization problem under a cardinality constraint. Its
optimal solution is to select the blocks with the largest positive gains
\(\delta_B\), up to the budget \(K\). Equivalently, one selects the top-\(K\)
blocks according to \(S_B=[\delta_B]_+\), omitting zero-score blocks if fewer
than \(K\) blocks have positive gain. Therefore, \(\Delta\)-Hessian selection is
optimal for minimizing the diagonal Hessian surrogate
\(\widetilde{L}(\mathcal{S})\).

We further relate this surrogate to the exact output-distortion objective. Let
\[
\mathbf{H}
=
\mathbf{D}
+
\mathbf{O},
\]
where \(\mathbf{O}\) contains the off-diagonal Hessian terms. For any selected
block set \(\mathcal{S}\),
\[
L(\mathcal{S})-\widetilde{L}(\mathcal{S})
=
\operatorname{Tr}
\left(
\mathbf{E}_{\mathcal{S}}
\mathbf{O}
\mathbf{E}_{\mathcal{S}}^\top
\right).
\]
Using the spectral norm bound,
\[
\left|
L(\mathcal{S})-\widetilde{L}(\mathcal{S})
\right|
\le
\|\mathbf{O}\|_2
\|\mathbf{E}_{\mathcal{S}}\|_F^2.
\]
Similarly, for the output-distortion reduction,
\[
\Delta(\mathcal{S})
=
L(\emptyset)-L(\mathcal{S}),
\qquad
\widetilde{\Delta}(\mathcal{S})
=
\widetilde{L}(\emptyset)-\widetilde{L}(\mathcal{S}),
\]
we have
\[
\left|
\Delta(\mathcal{S})-\widetilde{\Delta}(\mathcal{S})
\right|
\le
\|\mathbf{O}\|_2
\left(
\|\mathbf{R}\|_F^2
+
\|\mathbf{E}_{\mathcal{S}}\|_F^2
\right).
\]
Define
\[
\epsilon_K
=
\max_{|\mathcal{S}|\le K}
\left|
\Delta(\mathcal{S})-\widetilde{\Delta}(\mathcal{S})
\right|.
\]
Let \(\mathcal{S}^\star\) be the exact output-distortion oracle, i.e.,
\[
\mathcal{S}^\star
=
\arg\max_{|\mathcal{S}|\le K}
\Delta(\mathcal{S}),
\]
and let \(\widehat{\mathcal{S}}\) be the set selected by \(\Delta\)-Hessian,
i.e.,
\[
\widehat{\mathcal{S}}
=
\arg\max_{|\mathcal{S}|\le K}
\widetilde{\Delta}(\mathcal{S}).
\]
Then
\[
\Delta(\mathcal{S}^\star)
\le
\widetilde{\Delta}(\mathcal{S}^\star)+\epsilon_K
\le
\widetilde{\Delta}(\widehat{\mathcal{S}})+\epsilon_K
\le
\Delta(\widehat{\mathcal{S}})+2\epsilon_K.
\]
Therefore,
\[
\Delta(\mathcal{S}^\star)
-
\Delta(\widehat{\mathcal{S}})
\le
2\epsilon_K.
\]
This shows that the gap between the exact output-distortion oracle and
\(\Delta\)-Hessian selection is controlled by the approximation error between
the full Hessian objective and its diagonal surrogate. In particular, if the
off-diagonal Hessian term is zero, i.e., \(\mathbf{O}=0\), then
\(\epsilon_K=0\), and \(\Delta\)-Hessian selection exactly matches the
output-distortion oracle. When the off-diagonal interactions are small,
\(\Delta\)-Hessian provides a close block-wise approximation while allowing
efficient independent ranking.

\section{Benchmark Details}

\subsection{Baselines}\label{baseline}
We benchmark our methods compared with the following six baselines:
\begin{itemize}[leftmargin=1.2em]
    \item RTN~\citep{RTN} employs uniform rounding-to-nearest with fixed scaling to quantize weights and activations without calibration, thereby offering a almost zero-cost baseline at the expense of larger accuracy loss.
    \item GPTQ~\citep{gptq} employs post-training, Hessian-aware greedy weight quantization while applying error compensation to previously quantized rows, thereby preserving accuracy under W4A16.
    \item AWQ~\citep{awq_2024} employs activation-aware weight clipping and per-channel scaling while prioritizing salient channels, thereby achieving accurate W4A16 quantization with light calibration.
    \item SmoothQuant~\citep{smoothquant} employs channel-wise scaling to shift activation outliers into weights while smoothing activation distributions, thereby enabling W8A8 quantization with minimal degradation.
    \item QuaRot~\citep{quantization_quarot} employs online Hadamard-based orthogonal rotations to on-the-fly disperse outliers and quantize in the rotated space, thereby enabling W4A4 quantization with minimal overhead and small loss.
    \item SpinQuant~\citep{spinquant_quant} employs learned orthogonal rotations to homogenize weights and activations magnitudes while reducing outliers, thereby enabling W4A4 quantization with improved stability.
\end{itemize}

\section{Additional Results}

\subsection{Implementation Details}\label{implementation}
We perform MosaicQuant quantization on a single H20 with 96 GB of memory and evaluate it on the \emph{Workstation} platforms: an NVIDIA RTX~4090 (24~GB) GPU paired with an Intel Xeon Gold~6430 CPU and 120~GB DDR5 host memory. 
For W4A8 quantization, we use per-token dynamic activation quantization and per-channel weight quantization. For the 4-bit setting, we adopt per-group and per-channel symmetric quantization for activations and weights, respectively.

The smoothing factor $\boldsymbol{\lambda}\in\mathbb{R}^m$ is a per-channel vector whose $i$-th element is computed as $\lambda_i = \max(|\mathbf{X}_{:,i}|)^{\alpha}/\max(|\mathbf{W}_{i,:}|)^{1-\alpha}$ where $\mathbf{X}\in\mathbb{R}^{t\times m}$ and $\mathbf{W}\in\mathbb{R}^{m\times n}$ following SmoothQuant~\citep{smoothquant}. 
The migration strength $\alpha$ is chosen offline, per layer, by searching for the value that minimizes the layer output mean squared error (MSE) after MosaicQuant on the calibration dataset.

For the baselines, we run AWQ~\citep{awq_2024}, FlatQuant~\citep{flatquant}, QuaRot~\citep{quantization_quarot}, and QServe~\citep{quantizaion_qserve} using their publicly released implementations. 
For AWQ, FlatQuant, and QuaRot, we adopt the official PyTorch-based code and recommended W4A4/W4A8 configurations for each model, and ensure that all methods share the same decoding setup (greedy decoding), batch sizes, and sequence lengths on the same GPU. 
For QServe, we build its C++/CUDA runtime following the official instructions and evaluate the W4A8 configuration on the same hardware and workload as MosaicQuant. 
All throughput measurements are collected after warm-up runs to avoid initialization overhead.

\color{black}

\subsection{Zero-Shot Task Results}\label{zero_shot_appendix}
In \autoref{llama3} and \autoref{qwen3}, we show the complete results of \autoref{precision}. including ARC-Challenge, ARC-Easy~\citep{ARC-C}, HellaSwag~\citep{hellaswag}, LAMBADA~\citep{lambada}, PIQA\citep{PIQA}, and WinoGrande~\citep{WinoGrande}. Additionally, we also report the perplexity score on WikiText2~\citep{wiki_dataset} datasets. We compare our results with previous works including RTN~\citep{RTN}, GPTQ~\citep{gptq}, AWQ~\citep{awq_2024}, SmoothQuant~\citep{smoothquant}, QuaRot~\citep{quantization_quarot} and SpinQuant~\citep{spinquant_quant}.

\begin{table}[ht]
    \setlength{\tabcolsep}{3.4pt}
    \vspace{-15pt}
    \caption{
        Complete comparison of the perplexity score on WikiText2 and averaged accuracy on six zero-shot tasks on LLaMA3 3B and 8B. 
    }
    \label{llama3}
    \scriptsize \centering
    \begin{tabular}{ccccccccccc}
    \toprule
     \textbf{Model} & \textbf{Precision} & \textbf{Method} & \textbf{ARC-C} & \textbf{ARC-E} & \textbf{HellaSwag} & \textbf{PIQA} & \textbf{Winogrande} & \textbf{LAMBADA} & \textbf{Avg.} & \textbf{Wiki2} ($\downarrow$)\\
     
    \midrule
     \multirow{18}{*}{\makecell{3B}} 
     & W16A16 & -- & 47.6 & 69.9 & 71.0 & 76.0 & 66.6 & 65.9 & 66.2 & 10.7 \\
    \cmidrule{2-11}
     & W4A16 & RTN  & 41.3 & 60.2 & 66.2 & 73.1 & 63.0 & 61.6 & 60.9 & 18.8 \\
     & W4A16 & GPTQ & 41.4 & 60.8 & 65.9 & 73.6 & 65.0 & 63.4 & 61.7 & 15.2 \\
     & W4A16 & AWQ  & 43.5 & 66.7 & 65.8 & 75.8 & 62.7 & 63.8 & 63.1 & 12.7 \\
    \cmidrule{2-11}
     & W4A8 & RTN          & 42.6 & 60.2 & 66.2 & 72.6 & 62.7 & 60.1 & 60.7 & 29.0 \\
     & W4A8 & SmoothQuant  & 40.7 & 59.8 & 65.5 & 73.8 & 58.5 & 60.7 & 59.8 & 288.5 \\
     & W4A8 & QuaRot       & 46.9 & 67.3 & 68.1 & 75.6 & 64.8 & 62.2 & 64.2 & 12.4 \\
     & W4A8 & SpinQuant    & 47.2 & 68.8 & 69.2 & 76.0 & 66.7 & 63.2 & 65.2 & 11.5 \\
     & W4A8 & ResQ         & 44.4 & 69.4 & 70.2 & 75.2 & 65.2 & 64.2 & 64.8 & 11.8 \\
     & W4A8 & Atom         & 44.2 & 68.8 & 70.5 & 74.4 & 65.6 & 63.4 & 64.5 & 11.9 \\
     & \cellcolor{gray}W4A8 & \cellcolor{gray}MosaicQuant
        & \cellcolor{gray}46.8 & \cellcolor{gray}69.0 & \cellcolor{gray}70.2 & \cellcolor{gray}75.2
        & \cellcolor{gray}65.8 & \cellcolor{gray}64.8 & \cellcolor{gray}65.3 & \cellcolor{gray}11.3 \\
    \cmidrule{2-11}
     & W4A4 & RTN          & 29.8 & 41.0 & 41.4 & 57.3 & 50.9 & 38.9 & 43.2 & 741.9 \\
     & W4A4 & SmoothQuant  & 30.5 & 43.6 & 37.7 & 58.0 & 52.9 & 45.3 & 44.7 & 372.3 \\
     & W4A4 & QuaRot       & 42.3 & 64.9 & 64.8 & 72.9 & 63.8 & 59.8 & 61.4 & 16.9 \\
     & W4A4 & SpinQuant    & 44.9 & 68.3 & 67.2 & 75.0 & 65.5 & 63.5 & 64.1 & 11.9 \\
     & W4A4 & ResQ         & 43.1 & 66.6 & 68.4 & 74.1 & 63.2 & 62.8 & 63.0 & 12.2 \\
     & W4A4 & Atom         & 42.8 & 66.4 & 68.8 & 73.2 & 62.1 & 62.5 & 62.6 & 12.5 \\
     & \cellcolor{gray}W4A4 & \cellcolor{gray}MosaicQuant
        & \cellcolor{gray}45.8 & \cellcolor{gray}69.4 & \cellcolor{gray}69.8 & \cellcolor{gray}74.5
        & \cellcolor{gray}64.8 & \cellcolor{gray}64.8 & \cellcolor{gray}64.9 & \cellcolor{gray}11.8 \\

    \midrule
     \multirow{18}{*}{\makecell{8B}} 
     & W16A16 & -- & 54.86 & 79.55 & 79.13 & 80.74 & 73.72 & 72.93 & 73.5 & 8.64 \\
    \cmidrule{2-11}
     & W4A16 & RTN  & 52.5 & 77.7 & 77.4 & 79.6 & 73.4 & 66.7 & 71.2 & 10.5 \\
     & W4A16 & GPTQ & 54.2 & 79.4 & 78.1 & 80.1 & 72.1 & 70.6 & 72.4 & 9.0 \\
     & W4A16 & AWQ  & 53.8 & 77.6 & 78.8 & 80.0 & 73.2 & 72.1 & 72.6 & 10.3 \\
    \cmidrule{2-11}
     & W4A8 & RTN          & 52.7 & 77.0 & 77.3 & 79.4 & 73.0 & 66.3 & 71.0 & 10.6 \\
     & W4A8 & SmoothQuant  & 42.5 & 66.2 & 69.6 & 73.72 & 66.1 & 57.5 & 60.4 & 13.3 \\
     & W4A8 & QuaRot       & 53.4 & 78.0 & 77.9 & 79.7 & 71.0 & 69.5 & 71.6 & 10.6 \\
     & W4A8 & SpinQuant    & 54.0 & 78.5 & 78.1 & 79.6 & 72.4 & 71.3 & 72.3 & 10.3 \\
     & W4A8 & ResQ         & 54.2 & 78.9 & 78.1 & 78.6 & 71.8 & 71.5 & 72.2 & 10.4 \\
     & W4A8 & Atom         & 53.8 & 78.1 & 78.3 & 77.9 & 72.2 & 71.2 & 71.9 & 10.6 \\
     & \cellcolor{gray}W4A8 & \cellcolor{gray}MosaicQuant
        & \cellcolor{gray}54.3 & \cellcolor{gray}79.2 & \cellcolor{gray}78.5 & \cellcolor{gray}79.1
        & \cellcolor{gray}72.5 & \cellcolor{gray}72.3 & \cellcolor{gray}72.7 & \cellcolor{gray}10.2 \\
    \cmidrule{2-11}
     & W4A4 & RTN          & 28.3 & 40.2 & 43.8 & 59.1 & 50.0 & 21.0 & 40.4 & 92.9 \\
     & W4A4 & SmoothQuant  & 35.0 & 57.3 & 59.8 & 67.9 & 56.8 & 40.2 & 52.8 & 19.4 \\
     & W4A4 & QuaRot       & 49.2 & 73.4 & 74.1 & 77.4 & 68.3 & 65.8 & 68.0 & 13.1 \\
     & W4A4 & SpinQuant    & 51.9 & 75.2 & 75.9 & 77.5 & 70.8 & 68.4 & 70.0 & 11.4 \\
     & W4A4 & ResQ         & 51.6 & 76.5 & 75.2 & 76.3 & 70.5 & 69.4 & 69.9 & 11.6 \\
     & W4A4 & Atom         & 51.4 & 75.5 & 74.4 & 76.2 & 68.8 & 66.3 & 67.4 & 12.5 \\
     & \cellcolor{gray}W4A4 & \cellcolor{gray}MosaicQuant
        & \cellcolor{gray}52.7 & \cellcolor{gray}77.1 & \cellcolor{gray}76.3 & \cellcolor{gray}78.8
        & \cellcolor{gray}71.3 & \cellcolor{gray}70.8 & \cellcolor{gray}71.2 & \cellcolor{gray}10.8 \\
    \bottomrule
    \end{tabular}
    \vspace{-10pt}
\end{table}

\subsection{Kernel Profile}\label{kernel_profile}

Table~\ref{tab:env1_latency} shows that kernel fusion consistently reduces the latency of the main GEMM operators on Qwen3-8B. For attention, the fused $\text{qkv}_{\text{fused}}$ operator obtains $2.51$--$2.56\times$ prefill speedup and $1.61$--$1.79\times$ decode speedup across batch sizes, while $o_{\text{proj}}$ improves by $2.39$--$2.46\times$ in prefill and $1.45$--$1.87\times$ in decode. For MLP layers, fusing $\text{gate}/\text{up}_{\text{proj}}$ gives the largest prefill gain, reaching $2.64$--$2.68\times$, with $1.25$--$1.45\times$ decode speedup. The $\text{down}_{\text{proj}}$ operator shows a different pattern: its prefill speedup is $1.92$--$1.97\times$, but its decode speedup increases with batch size and reaches $2.90\times$ at batch size 32. Overall, these results show that kernel fusion is effective across both attention and MLP projections, with particularly strong gains in prefill and for decode in $\text{down}_{\text{proj}}$.

{\scriptsize
\setlength{\tabcolsep}{3.4pt}
\renewcommand{\arraystretch}{0.95}

\begin{longtable}{ccccccccccc}
\caption{Complete comparison of the perplexity score on WikiText2 and averaged accuracy on six zero-shot tasks on Qwen3 4B/8B/14B/32B.}
\label{qwen3}\\

\toprule
\textbf{Model} & \textbf{Precision} & \textbf{Method} & \textbf{ARC-C} & \textbf{ARC-E} & \textbf{HellaSwag} & \textbf{PIQA} & \textbf{Winogrande} & \textbf{LAMBADA} & \textbf{Avg.} & \textbf{Wiki2} ($\downarrow$)\\
\midrule
\endfirsthead

\multicolumn{11}{l}{\scriptsize\itshape Table~\ref{qwen3} -- continued from previous page}\\
\toprule
\textbf{Model} & \textbf{Precision} & \textbf{Method} & \textbf{ARC-C} & \textbf{ARC-E} & \textbf{HellaSwag} & \textbf{PIQA} & \textbf{Winogrande} & \textbf{LAMBADA} & \textbf{Avg.} & \textbf{Wiki2} ($\downarrow$)\\
\midrule
\endhead

\midrule
\multicolumn{11}{r}{\scriptsize\itshape Continued on next page}\\
\endfoot

\bottomrule
\endlastfoot

\multirow{18}{*}{\makecell{4B}} 
& W16A16 & -- & 58.6 & 81.1 & 69.09 & 76.0 & 68.11 & 63.56 & 69.4 & 10.04 \\
\cmidrule{2-11}
& W4A16 & RTN  & 56.7 & 77.3 & 66.8 & 73.2 & 66.8 & 61.9 & 67.1 & 10.6 \\
& W4A16 & GPTQ & 58.1 & 80.1 & 68.4 & 75.6 & 67.0 & 62.1 & 68.6 & 10.4 \\
& W4A16 & AWQ  & 57.2 & 80.7 & 68.8 & 75.8 & 66.1 & 61.5 & 68.4 & 10.4 \\
\cmidrule{2-11}
& W4A8 & RTN          & 42.3 & 68.7 & 64.1 & 68.8 & 53.9 & 55.1 & 58.8 & 30.6 \\
& W4A8 & SmoothQuant  & 42.2 & 69.5 & 63.4 & 71.5 & 57.5 & 55.0 & 59.9 & 22.6 \\
& W4A8 & QuaRot       & 55.3 & 79.5 & 68.3 & 74.7 & 67.2 & 61.8 & 67.8 & 11.2 \\
& W4A8 & SpinQuant    & 56.2 & 80.4 & 68.2 & 75.5 & 67.6 & 62.4 & 68.4 & 10.8 \\
& W4A8 & ResQ         & 56.4 & 80.7 & 67.8 & 75.2 & 67.8 & 62.6 & 68.4 & 10.7 \\
& W4A8 & Atom         & 56.0 & 79.8 & 67.4 & 74.8 & 66.9 & 62.1 & 67.8 & 11.0 \\
& \cellcolor{gray}W4A8 & \cellcolor{gray}MosaicQuant
  & \cellcolor{gray}57.6 & \cellcolor{gray}80.5 & \cellcolor{gray}68.2 & \cellcolor{gray}75.6
  & \cellcolor{gray}67.6 & \cellcolor{gray}62.8 & \cellcolor{gray}68.7 & \cellcolor{gray}10.4 \\
\cmidrule{2-11}
& W4A4 & RTN          & 26.7 & 28.6 & 44.2 & 48.9 & 51.8 & 48.9 & 41.5 & 8791 \\
& W4A4 & SmoothQuant  & 22.6 & 25.9 & 42.3 & 51.6 & 47.9 & 50.2 & 40.1 & 9910 \\
& W4A4 & QuaRot       & 54.93 & 70.58 & 59.07 & 73.29 & 58.72 & 53.46 & 61.7 & 12.5 \\
& W4A4 & SpinQuant    & 55.4 & 76.8 & 66.5 & 74.8 & 63.5 & 56.7 & 65.6 & 11.5 \\
& W4A4 & ResQ         & 54.5 & 76.8 & 66.8 & 73.4 & 62.8 & 56.5 & 65.1 & 12.0 \\
& W4A4 & Atom         & 54.8 & 76.2 & 66.9 & 73.0 & 62.5 & 55.4 & 64.8 & 12.3 \\
& \cellcolor{gray}W4A4 & \cellcolor{gray}MosaicQuant
  & \cellcolor{gray}56.4 & \cellcolor{gray}78.54 & \cellcolor{gray}67.08 & \cellcolor{gray}74.27
  & \cellcolor{gray}63.85 & \cellcolor{gray}57.89 & \cellcolor{gray}66.3 & \cellcolor{gray}10.96 \\
\midrule

\multirow{18}{*}{\makecell{8B}} 
& W16A16 & -- & 55.5 & 83.5 & 78.8 & 76.4 & 68.0 & 67.4 & 71.6 & 9.71 \\
\cmidrule{2-11}
& W4A16 & RTN  & 53.8 & 79.1 & 76.3 & 75.4 & 63.6 & 65.6 & 69.0 & 12.0 \\
& W4A16 & GPTQ & 55.6 & 80.1 & 77.8 & 76.0 & 66.1 & 66.8 & 70.4 & 10.8 \\
& W4A16 & AWQ  & 54.7 & 83.1 & 77.1 & 75.4 & 67.5 & 66.7 & 70.8 & 10.2 \\
\cmidrule{2-11}
& W4A8 & RTN          & 45.8 & 73.4 & 72.3 & 73.1 & 62.9 & 58.7 & 64.4 & 12.3 \\
& W4A8 & SmoothQuant  & 44.3 & 71.0 & 71.7 & 73.4 & 62.1 & 60.0 & 63.8 & 12.5 \\
& W4A8 & QuaRot       & 53.9 & 77.8 & 74.8 & 72.7 & 62.6 & 60.5 & 67.1 & 11.5 \\
& W4A8 & SpinQuant    & 54.2 & 81.9 & 77.5 & 75.6 & 66.5 & 66.8 & 70.4 & 10.8 \\
& W4A8 & ResQ         & 54.5 & 81.8 & 77.4 & 75.6 & 66.2 & 66.4 & 70.3 & 10.7 \\
& W4A8 & Atom         & 55.0 & 82.5 & 76.5 & 75.2 & 66.8 & 65.8 & 70.3 & 10.9 \\
& \cellcolor{gray}W4A8 & \cellcolor{gray}MosaicQuant
  & \cellcolor{gray}54.8 & \cellcolor{gray}82.8 & \cellcolor{gray}77.9 & \cellcolor{gray}75.7
  & \cellcolor{gray}67.0 & \cellcolor{gray}65.6 & \cellcolor{gray}70.6 & \cellcolor{gray}10.6 \\
\cmidrule{2-11}
& W4A4 & RTN          & 22.6 & 24.9 & 45.6 & 48.9 & 51.8 & 46.9 & 40.1 & 4392 \\
& W4A4 & SmoothQuant  & 25.7 & 25.5 & 41.7 & 50.5 & 52.2 & 44.9 & 40.1 & 3360.1 \\
& W4A4 & QuaRot       & 49.8 & 77.8 & 69.8 & 74.3 & 65.7 & 60.1 & 66.3 & 15.6 \\
& W4A4 & SpinQuant    & 53.6 & 78.5 & 71.4 & 76.5 & 67.3 & 62.4 & 68.3 & 11.4 \\
& W4A4 & ResQ         & 52.1 & 78.8 & 73.8 & 71.5 & 63.5 & 61.1 & 66.8 & 14.2 \\
& W4A4 & Atom         & 51.8 & 76.6 & 72.6 & 71.2 & 62.8 & 61.5 & 66.1 & 14.5 \\
& \cellcolor{gray}W4A4 & \cellcolor{gray}MosaicQuant
  & \cellcolor{gray}53.8 & \cellcolor{gray}80.2 & \cellcolor{gray}75.4 & \cellcolor{gray}72.8
  & \cellcolor{gray}65.3 & \cellcolor{gray}63.8 & \cellcolor{gray}68.6 & \cellcolor{gray}11.2 \\
\midrule

\multirow{18}{*}{\makecell{14B}} 
& W16A16 & -- & 59.0 & 84.3 & 80.5 & 80.0 & 72.9 & 68.4 & 74.2 & 8.6 \\
\cmidrule{2-11}
& W4A16 & RTN  & 51.8 & 80.6 & 78.5 & 77.9 & 68.7 & 64.4 & 70.3 & 9.9 \\
& W4A16 & GPTQ & 57.1 & 81.9 & 78.8 & 78.8 & 72.5 & 66.9 & 72.7 & 9.2 \\
& W4A16 & AWQ  & 51.7 & 79.2 & 79.3 & 76.0 & 66.1 & 66.5 & 69.8 & 9.6 \\
\cmidrule{2-11}
& W4A8 & RTN          & 50.7 & 79.5 & 75.4 & 75.3 & 66.9 & 62.0 & 68.3 & 10.9 \\
& W4A8 & SmoothQuant  & 50.5 & 79.0 & 75.8 & 75.8 & 66.3 & 61.3 & 68.1 & 11.0 \\
& W4A8 & QuaRot       & 56.4 & 80.7 & 76.1 & 76.9 & 68.5 & 65.8 & 70.7 & 10.4 \\
& W4A8 & SpinQuant    & 57.8 & 81.8 & 77.6 & 78.3 & 71.8 & 67.6 & 72.5 & 9.8 \\
& W4A8 & ResQ         & 57.2 & 82.5 & 78.4 & 77.5 & 72.0 & 66.6 & 72.4 & 12.1 \\
& W4A8 & Atom         & 57.5 & 81.4 & 78.2 & 77.7 & 71.5 & 66.2 & 72.1 & 12.4 \\
& \cellcolor{gray}W4A8 & \cellcolor{gray}MosaicQuant
  & \cellcolor{gray}58.6 & \cellcolor{gray}83.1 & \cellcolor{gray}79.3 & \cellcolor{gray}79.2
  & \cellcolor{gray}72.4 & \cellcolor{gray}68.0 & \cellcolor{gray}73.4 & \cellcolor{gray}11.5 \\
\cmidrule{2-11}
& W4A4 & RTN          & 24.8 & 23.9 & 50.6 & 55.7 & 46.8 & 50.7 & 42.1 & 18749 \\
& W4A4 & SmoothQuant  & 26.5 & 25.8 & 48.7 & 51.2 & 50.2 & 45.3 & 41.3 & 21675 \\
& W4A4 & QuaRot       & 52.8 & 76.4 & 72.4 & 73.6 & 65.0 & 62.8 & 67.2 & 18.2 \\
& W4A4 & SpinQuant    & 56.3 & 81.0 & 77.8 & 76.9 & 71.4 & 66.7 & 71.7 & 12.0 \\
& W4A4 & ResQ         & 56.7 & 80.5 & 76.4 & 76.1 & 70.8 & 65.6 & 71.0 & 12.8 \\
& W4A4 & Atom         & 56.1 & 80.2 & 75.5 & 75.2 & 70.2 & 65.7 & 70.5 & 14.5 \\
& \cellcolor{gray}W4A4 & \cellcolor{gray}MosaicQuant
  & \cellcolor{gray}57.9 & \cellcolor{gray}82.4 & \cellcolor{gray}78.9 & \cellcolor{gray}79.2
  & \cellcolor{gray}72.4 & \cellcolor{gray}66.5 & \cellcolor{gray}72.9 & \cellcolor{gray}11.8 \\
\midrule

\multirow{18}{*}{\makecell{32B}} 
& W16A16 & -- & 57.8 & 84.4 & 84.2 & 80.9 & 73.6 & 70.3 & 75.2 & 7.6 \\
\cmidrule{2-11}
& W4A16 & RTN  & 49.7 & 73.0 & 82.6 & 71.9 & 62.9 & 68.5 & 68.1 & 38.5 \\
& W4A16 & GPTQ & 57.8 & 82.6 & 83.6 & 79.7 & 67.6 & 69.8 & 73.5 & 8.3 \\
& W4A16 & AWQ  & 56.9 & 82.5 & 83.2 & 79.8 & 71.8 & 68.7 & 73.8 & 8.2 \\
\cmidrule{2-11}
& W4A8 & RTN          & 49.5 & 76.3 & 78.6 & 72.9 & 65.3 & 62.0 & 67.4 & 11.2 \\
& W4A8 & SmoothQuant  & 49.2 & 75.8 & 78.9 & 71.4 & 63.5 & 63.7 & 67.1 & 11.6 \\
& W4A8 & QuaRot       & 51.3 & 78.9 & 79.8 & 76.8 & 70.9 & 68.5 & 71.0 & 10.5 \\
& W4A8 & SpinQuant    & 56.6 & 82.9 & 80.4 & 79.4 & 72.0 & 67.1 & 73.1 & 8.9 \\
& W4A8 & ResQ         & 55.7 & 82.9 & 81.2 & 79.6 & 71.8 & 66.8 & 73.0 & 9.2 \\
& W4A8 & Atom         & 55.4 & 82.3 & 80.9 & 80.4 & 72.4 & 66.6 & 73.0 & 9.3 \\
& \cellcolor{gray}W4A8 & \cellcolor{gray}MosaicQuant
  & \cellcolor{gray}56.9 & \cellcolor{gray}82.8 & \cellcolor{gray}83.5 & \cellcolor{gray}80.1
  & \cellcolor{gray}73.2 & \cellcolor{gray}68.9 & \cellcolor{gray}74.2 & \cellcolor{gray}8.4 \\
\cmidrule{2-11}
& W4A4 & RTN          & 28.5 & 27.6 & 60.8 & 52.5 & 50.7 & 52.0 & 45.4 & 1796 \\
& W4A4 & SmoothQuant  & 26.0 & 28.9 & 56.7 & 51.9 & 54.1 & 48.6 & 44.4 & 1806 \\
& W4A4 & QuaRot       & 49.8 & 72.2 & 75.6 & 73.4 & 67.9 & 65.3 & 67.4 & 15.6 \\
& W4A4 & SpinQuant    & 56.7 & 81.6 & 81.2 & 78.9 & 71.8 & 65.8 & 72.7 & 10.3 \\
& W4A4 & ResQ         & 56.9 & 81.6 & 80.1 & 78.4 & 71.2 & 63.5 & 72.0 & 10.9 \\
& W4A4 & Atom         & 57.5 & 81.2 & 80.6 & 78.2 & 71.5 & 63.9 & 72.2 & 10.5 \\
& \cellcolor{gray}W4A4 & \cellcolor{gray}MosaicQuant
  & \cellcolor{gray}57.2 & \cellcolor{gray}82.8 & \cellcolor{gray}82.4 & \cellcolor{gray}79.5
  & \cellcolor{gray}72.7 & \cellcolor{gray}65.0 & \cellcolor{gray}73.3 & \cellcolor{gray}8.8 \\
\end{longtable}
}

\begin{table}[t]
\centering
\small
\setlength{\tabcolsep}{4pt}
\renewcommand{\arraystretch}{1.2}
\resizebox{\textwidth}{!}{%
\begin{tabular}{cccccccc}
\toprule
\multirow{2}{*}{GEMM Operator} & \multirow{2}{*}{Batch Size}
& \multicolumn{2}{c}{without Kernel Fusion}
& \multicolumn{2}{c}{with Kernel Fusion}
& \multicolumn{2}{c}{Speedup} \\
\cmidrule(lr){3-4}\cmidrule(lr){5-6}\cmidrule(lr){7-8}
& & \makecell{Prefill\\(ms)} & \makecell{Decode\\(ms)}
& \makecell{Prefill\\(ms)} & \makecell{Decode\\(ms)}
& Prefill & Decode \\
\midrule
\multirow{4}{*}{\makecell{$\text{qkv}_{\text{fused}}$\\(Fused $q/k/v_{\text{proj}}$)\\$6144\times4096$}}
& 4  & 4.536  & 0.067 & 1.806  & 0.037 & 2.51$\times$ & 1.79$\times$ \\
& 8  & 9.070  & 0.066 & 3.581  & 0.038 & 2.53$\times$ & 1.74$\times$ \\
& 16 & 18.167 & 0.066 & 7.101  & 0.038 & 2.56$\times$ & 1.74$\times$ \\
& 32 & 36.344 & 0.062 & 14.270 & 0.038 & 2.55$\times$ & 1.61$\times$ \\
\midrule
\multirow{4}{*}{\makecell{$o_{\text{proj}}$\\$4096\times4096$}}
& 4  & 3.123  & 0.066 & 1.307  & 0.035 & 2.39$\times$ & 1.87$\times$ \\
& 8  & 6.301  & 0.065 & 2.624  & 0.038 & 2.40$\times$ & 1.73$\times$ \\
& 16 & 12.538 & 0.055 & 5.094  & 0.038 & 2.46$\times$ & 1.45$\times$ \\
& 32 & 25.043 & 0.062 & 10.228 & 0.038 & 2.45$\times$ & 1.63$\times$ \\
\midrule
\multirow{4}{*}{\makecell{$\text{gate\_up}_{\text{fused}}$\\(Fused $\text{gate}/\text{up}_{\text{proj}}$)\\$24576\times4096$}}
& 4  & 17.171  & 0.076 & 6.498  & 0.060 & 2.64$\times$ & 1.26$\times$ \\
& 8  & 34.343  & 0.077 & 12.835 & 0.061 & 2.68$\times$ & 1.25$\times$ \\
& 16 & 68.780  & 0.082 & 25.949 & 0.064 & 2.65$\times$ & 1.29$\times$ \\
& 32 & 137.811 & 0.097 & 52.298 & 0.067 & 2.64$\times$ & 1.45$\times$ \\
\midrule
\multirow{4}{*}{\makecell{$\text{down}_{\text{proj}}$\\$4096\times12288$}}
& 4  & 9.922  & 0.148 & 5.161  & 0.077 & 1.92$\times$ & 1.93$\times$ \\
& 8  & 19.824 & 0.148 & 10.141 & 0.059 & 1.95$\times$ & 2.49$\times$ \\
& 16 & 39.621 & 0.158 & 20.174 & 0.060 & 1.96$\times$ & 2.63$\times$ \\
& 32 & 79.616 & 0.180 & 40.376 & 0.062 & 1.97$\times$ & 2.90$\times$ \\
\bottomrule
\end{tabular}%
}
\caption{Latency comparison with/without kernel fusion on Qwen3-8B in Environment 1. The fused operators $\text{qkv}_{\text{fused}}$ and $\text{gate\_up}_{\text{fused}}$ merge the original $q/k/v_{\text{proj}}$ and $\text{gate}/\text{up}_{\text{proj}}$ into a single GEMM with concatenated output dimensions.}
\label{tab:env1_latency}
\end{table}

\end{document}